\documentclass{article}
\usepackage[final]{neurips_2025}

\usepackage[utf8]{inputenc} 
\usepackage{CJKutf8} 
\usepackage[T1]{fontenc}    
\usepackage{hyperref}       
\usepackage{url}            
\usepackage{booktabs}       
\usepackage{amsfonts}       
\usepackage{amssymb}        
\usepackage{nicefrac}       
\usepackage{microtype}      
\usepackage{xcolor}         
\usepackage{times}
\usepackage{latexsym}
\usepackage{microtype}
\usepackage{inconsolata}
\usepackage{multirow}
\usepackage[spacesep,definevectors]{easyvector}
\usepackage{amsmath}
\usepackage{stmaryrd}       
\usepackage{mathtools}
\usepackage{xspace}
\usepackage{algpseudocode}
\usepackage{algorithm}
\usepackage{tikz}
\usepackage{hyperref}
\usepackage{CJKutf8}
\usepackage{enumitem}
\usepackage{multirow}
\usepackage{graphicx}
\usepackage{booktabs}
\usepackage{multirow}
\usepackage{graphicx}
\usepackage{colortbl}
\usepackage{xcolor}
\usepackage{url}
\usepackage{wrapfig}
\usepackage{amssymb}
\usepackage{pifont}
\usepackage{tikz}
\usepackage{pgfplots}
\pgfplotsset{compat=1.18}
\usepackage{pgfplotstable}
\usepackage{sansmath}
\usepackage{booktabs}
\usepackage{adjustbox}
\usepackage{array}
\usepackage{amsthm,amsmath,amssymb}

\usepackage{booktabs}
\usepackage{multirow}
\usepackage{array}
\usepackage{tabularx}

\usepackage{xcolor}
\usepackage{mdframed}
\usepackage{fvextra}

\newmdenv[
  linewidth=0.8pt,
  linecolor=black,
  backgroundcolor=gray!5,
  roundcorner=4pt,
  innertopmargin=8pt,
  innerbottommargin=8pt,
  innerleftmargin=8pt,
  innerrightmargin=8pt,
  skipabove=8pt,
  skipbelow=8pt
]{promptbox}

\newtheorem{definition}{Definition}

\definecolor{mtag}{RGB}{153,204,255}
\definecolor{atag}{RGB}{255,191,179}
\definecolor{gtag}{RGB}{218,200,235}
\definecolor{ttag}{RGB}{239,222,180}

\newcommand{\OurMODEL}{\textbf{\textsc{Spring}}} 

\newcommand{\eat}[1]{}

\title{Rewarding Novel Deductions: Solver-guided Process Supervision for Logical Reasoning}

\author{
Muhammad Asif Ali\textsuperscript{1,2},
Wenqing Wang\textsuperscript{3},
Huan Wang\textsuperscript{3},
Mohammad Raza\textsuperscript{4,\textdagger}
\\[0.6em]
\textsuperscript{1}FORTE Lab,
\textsuperscript{2}Faculty of Science,
Information Technology University,
Lahore, Pakistan
\\
\textsuperscript{3}College of Informatics,
Huazhong Agricultural University,
Wuhan, China
\\
\textsuperscript{4}Qatar Computing Research Institute,
Hamad Bin Khalifa University,
Doha, Qatar
\\[0.4em]
\textsuperscript{\textdagger}\textbf{Corresponding author:}
\textcolor{blue}{\texttt{mraza@hbku.edu.qa}}
}

\begin{document}

\maketitle

\begin{abstract}
Logical reasoning remains a major challenge for large language models (LLMs), particularly on structured problems that require precise constraint tracking, consistency preservation, and multi-step deduction. This challenge is especially acute for small-scale LLMs, which are more prone to producing inconsistent, redundant, or brittle reasoning trajectories.
Existing approaches for improving logical reasoning largely optimize for final-answer correctness, providing only weak supervision over the intermediate reasoning process. 
In this work, we propose~\OurMODEL{}: (\textbf{\underline{S}}olver-guided \textbf{\underline{P}}rocess \textbf{\underline{R}}ewards for Novel Log\textbf{\underline{I}}cal Reaso\textbf{\underline{N}}ing Step \textbf{\underline{G}}eneration). 
\OurMODEL{} uses an SMT solver as a training-time verifier of intermediate reasoning steps to provide process-level supervision.
It introduces the notion of a \emph{novel reasoning step}, namely, a step that is logically valid, consistent with the evolving reasoning state, and not already implied by previously accepted non-contradictory deductions. 
Based on this solver-based assessment, it designs process rewards that encourage novel inferential progress while penalizing contradictory and uninformative reasoning steps.
Evaluation across three logical reasoning benchmarks: ZebraLogic, AR-LSAT, and Knights and Knaves, and four LLMs shows that~\OurMODEL{} consistently outperforms base LLMs, outcome-only reward baselines, and Logic-LM. On ZebraLogic, SPRING improves puzzle accuracy by up to 49.71 and 15.43 points over the base LLM and strongest outcome-only baseline, respectively. On AR-LSAT, it improves overall accuracy by up to 64.43 and 11.39 percentage points over the base LLM and strongest
outcome-only baseline, respectively. On Knights and Knaves,~\OurMODEL{} achieves up to 93.14 puzzle accuracy and 96.05 person accuracy.

\end{abstract}

\vspace{-0.7ex}
\section{Introduction}
\label{sec:intro}
\vspace{-0.7ex}
Logical reasoning is a fundamental capability for intelligent systems because it requires deriving conclusions that remain consistent with a set of premises and constraints. Unlike open-ended generation tasks, logical reasoning problems such as relational puzzles, symbolic deduction, and structured constraint-satisfaction tasks require not only a correct final answer, but also a sequence of intermediate deductions that is logically sound and progressively informative. Although LLMs have shown strong performance on a wide range of reasoning benchmarks, they still struggle with tasks that require precise deduction, consistency tracking, and long-horizon constraint maintenance~\citep{wei2022chain,kojima2022large,mirzadeh2024gsm}. Their reasoning chains are often redundant, unsupported by the premises, or internally contradictory, even when the final answer appears plausible~\citep{lightman2023let,cobbe2021training,mirzadeh2024gsm}. This limitation is especially pronounced for \emph{small-scale} LLMs, particularly models in the 4B-and-below regime, which often require stronger or more structured supervision to maintain coherent multi-step reasoning trajectories~\citep{li2025small,zhang2025making}.
While such models are attractive because of their lower computational cost and easier deployment, they are more prone to breakdowns in multi-step logical reasoning and often struggle to maintain coherent reasoning traces over long deduction chains~\citep{wei2022chain,li2025small,zhang2025making}.

Motivated by these limitations, a growing line of work augments LLM reasoning with symbolic solvers such as SAT, SMT, Z3, and theorem provers, which provide exact checks for satisfiability, implication, and consistency~\citep{barrett2018satisfiability,de2008z3}. Recent neurosymbolic methods improve reliability by translating natural-language problems into executable formal representations and delegating inference or verification to the solver~\citep{ye2023satlm,pan2023logiclm,olausson2023linc,hu2025ltrag}. More recent works have also begun exploring the use of formal solvers and theorem provers for process supervision during training, providing feedback on the correctness of intermediate reasoning steps~\citep{conf/iclr/KimY26}. However, logical correctness alone does not guarantee meaningful reasoning progress. An LLM may repeatedly restate previously established facts or generate trivial variations of existing deductions to produce logically valid but redundant steps that do not make any inferential progress. Effective process supervision must therefore go beyond verifying the validity of individual deductions to also assess whether they contribute new information to the evolving reasoning state and make genuine progress towards a solution.

\eat{
To address this, we introduce~\OurMODEL{}: (\textbf{\underline{S}}olver-guided \textbf{\underline{P}}rocess \textbf{\underline{R}}ewards for Novel Log\textbf{\underline{I}}cal Reaso\textbf{\underline{N}}ing Step \textbf{\underline{G}}eneration), a solver-guided reinforcement learning framework for logical reasoning. 
Unlike prior solver-augmented methods that mainly use symbolic solvers for final inference or verification,~\OurMODEL{} uses the solver as a \emph{training-time process supervisor}. \OurMODEL{} introduces the notion of a \emph{novel reasoning step} (Definition~\ref{def:novel-step}), defined as a step that is logically valid, consistent with the evolving reasoning state, and not already implied by previously accepted non-contradictory deductions. 
During reinforcement learning, the solver evaluates the intermediate reasoning trajectory to identify novel steps and provide process-level feedback, which we then convert into rewards that explicitly encourage novel and logically useful reasoning steps while penalizing contradictory steps (Definition~\ref{def:contradict-step}), redundancy, and other uninformative deductions.}

To address this challenge, we introduce~\OurMODEL{} (\textbf{\underline{S}}olver-guided \textbf{\underline{P}}rocess \textbf{\underline{R}}ewards for Novel Log\textbf{\underline{I}}cal Reaso\textbf{\underline{N}}ing Step \textbf{\underline{G}}eneration), a reinforcement learning framework that uses symbolic solvers to explicitly reward inferential progress during logical reasoning. Central to \OurMODEL{} is the notion of rewarding \emph{novel reasoning steps}. A key question is how one  distinguishes genuinely novel deductions from redundant ones, since both must be logically implied by the problem premises. We address this by distinguishing between deductions that follow from the premises and those that are already implied by previously established reasoning steps alone: a deduction is considered novel if it is logically entailed by the premises but not by the preceding deductions in isolation. This distinction provides a principled basis for rewarding reasoning that progressively extends what has already been established. During reinforcement learning, a symbolic solver evaluates intermediate deductions to provide fine-grained process feedback, rewarding novel, logically supported deductions while penalizing contradictions and discouraging redundant or uninformative steps.

We evaluate~\OurMODEL{} across three logical reasoning benchmarks, ZebraLogic,
AR-LSAT, and Knights and Knaves, using four LLMs spanning different model
sizes and reasoning capabilities. Experimental results show that~\OurMODEL{}
consistently outperforms base LLMs, outcome-only reward baselines, and
Logic-LM. On ZebraLogic,~\OurMODEL{} improves puzzle accuracy by up to 49.71
points over the base LLM and 15.43 points over the strongest outcome-only
baseline. On AR-LSAT,~\OurMODEL{} improves overall accuracy by up to 64.43 and
11.39 points over the base LLM and strongest outcome-only baseline,
respectively. On Knights and Knaves,~\OurMODEL{} achieves up to 93.14 puzzle
accuracy and 96.05 person accuracy. Beyond final-task performance, our
analysis shows that~\OurMODEL{} produces shorter and cleaner successful traces,
fewer contradictions, and reasoning trajectories that are more consistent
with the underlying logical constraints. We summarize the key contributions
of this work as follows:

\begin{itemize}[noitemsep, topsep=2pt]
    \item  We formulate logical reasoning training as a process-supervised reinforcement learning problem that explicitly rewards inferential progress, rather than relying solely on final-answer correctness or intermediate-step validity.

    \item We introduce the notion of novel reasoning steps and show how a
    symbolic solver can verify them during training, thereby providing
    explicit supervision for inferential progress, validity, and logical
    consistency.

    \item We propose~\OurMODEL{}, a solver-guided reward framework that improves
    both reasoning trajectories and final logical reasoning performance by
    rewarding novel deductions while penalizing contradictory and
    uninformative reasoning steps.

    \item We conduct comprehensive experiments across three logical reasoning
    benchmarks and four LLMs, showing that~\OurMODEL{} consistently outperforms
    base LLMs, outcome-only reward baselines, and Logic-LM. Our analyses
    further show that~\OurMODEL{} produces shorter, cleaner, and more
    solver-consistent reasoning trajectories.
\end{itemize}

\vspace{-0.7ex}
\section{Related Work}
\label{sec:literature}
\vspace{-0.7ex}
\noindent{\bf Solver-Augmented Logical Reasoning for LLMs.}
A prominent line of recent work improves logical reasoning in LLMs by coupling them with external symbolic solvers rather than relying on unconstrained natural-language reasoning alone. In these approaches, the LLM typically serves as a semantic parser or translator that maps natural-language premises into a formal representation, while a symbolic backend performs exact deductive inference or consistency checking. This paradigm has become especially important for structured logical reasoning tasks, where global constraints, exclusivity conditions, and combinatorial dependencies are difficult to maintain through free-form chain-of-thought alone. Benchmark development has reinforced this need: AR-LSAT, introduced in \emph{Analytical Reasoning of Text}, remains a challenging dataset for structured logical deduction, while more recent puzzle-style benchmarks such as ZebraLogic expose substantial performance drops as deductive complexity increases \citep{zhong2022analytical,lin2025zebralogic}. 

Among solver-integrated methods, SatLM formulates reasoning problems declaratively, using the LLM to generate a satisfiability specification that is then solved by an automated theorem prover or SAT-style backend \citep{ye2023satlm}. Logic-LM further develops this paradigm by translating natural-language reasoning problems into symbolic programs, invoking a deterministic solver, and then using self-refinement to improve formalization quality; importantly, it reports gains on multiple logical reasoning benchmarks including AR-LSAT \citep{pan2023logiclm}. LINC follows a similar neurosymbolic design, but grounds reasoning in first-order logic and theorem proving, showing that explicit logical formalization can substantially improve deductive reliability over purely textual reasoning \citep{olausson2023linc}. More recently, LTRAG improves this general family of methods by enhancing autoformalization and self-refinement with retrieval-augmented exemplars, and reports further gains over Logic-LM and LINC on FOLIO and AR-LSAT \citep{hu2025ltrag}. Collectively, these works show that symbolic solvers can significantly augment LLMs on logical reasoning tasks when the model is able to produce executable formal representations. 

At the same time, recent studies show that the benefit of solver augmentation depends not only on whether a solver is used, but also on \emph{how} it is used and how well the LLM can interface with the solver’s symbolic language.
\citet{lam2024closer} perform a controlled comparison across multiple symbolic tools, including Z3, Pyke, and Prover9, and show that executable translation quality varies substantially across tools and is strongly tied to downstream reasoning accuracy. This observation is important because it clarifies that many current solver-augmented pipelines are bottlenecked by formalization quality: if the LLM fails to generate a faithful executable representation, the solver cannot recover the correct reasoning process. Thus, much of the literature on solver-augmented logical reasoning has focused on using solvers as \emph{final inference engines}, \emph{answer verifiers}, or \emph{self-refinement signals} over formalized problems. More recently, formal solvers and theorem provers have also been explored as sources of process supervision during training, providing feedback on the correctness of intermediate reasoning steps~\citep{conf/iclr/KimY26}.

\eat{This distinction is especially relevant for logic problems such as Zebra-style puzzles and AR-LSAT, where success depends not only on arriving at a satisfiable final solution, but also on maintaining a sequence of intermediate deductions that are informative, non-redundant, and contradiction-free. Prior solver-augmented methods clearly demonstrate the value of symbolic reasoning backends for these tasks, but they largely optimize the \emph{end product} of formalization and solving.} 

\noindent{\bf Distinction from Existing Solver-Augmented Methods.}
While formal verification can establish the logical correctness of intermediate reasoning steps, correctness alone does not ensure that a reasoning trajectory makes meaningful progress. A model may generate deductions that are logically valid but merely restate or reformulate information already established in earlier steps. \OurMODEL{} addresses this limitation by explicitly incorporating \emph{inferential novelty} into solver-guided process supervision rather than rewarding deductions solely for their logical validity. In this way, \OurMODEL{} extends solver-based process supervision beyond correctness verification to explicitly reward the generation of new logical information.

\vspace{-0.7ex}
\section{Problem Formulation}
\label{sec:problem}
\vspace{-0.7ex}
Let $x$ denote a logic problem instance consisting of (i) a set of natural-language premises and/or clues $\mathcal{C}$ and (ii) a set of domain constraints $\Gamma_{\mathrm{dom}}$ encoding the structural rules of the problem, such as exclusivity, uniqueness, admissible assignments, or variable ranges. We define the \emph{problem theory} as $\Gamma_{\mathrm{base}} = \Gamma_{\mathrm{dom}} \cup \mathcal{C}$. Given an input problem $x$, our goal is to train a language model policy $\pi_{\theta}$ to generate a reasoning trajectory $\tau = (s_1, s_2, \dots, s_T)$ followed by a final solution $y$, where each step $s_t$ is \emph{interleaved} and contains both a natural-language explanation and a formal logical statement executable in Z3. Concretely, each step is represented as $s_t = (u_t, z_t)$, where $u_t$ is a free-form natural-language rationale and $z_t$ is an optional formal statement in a constrained logical language. The full model output is therefore $o = (\tau, y) \sim \pi_{\theta}(\cdot \mid x)$. Our objective is not only to predict the correct final solution, but to learn a policy that produces \emph{reasoning trajectories} whose intermediate formal steps are both \emph{valid} and \emph{novel}. Intuitively, a valid step is one that is logically supported by the underlying problem theory, while a novel step contributes new information beyond previously generated non-contradictory steps and does not merely restate an existing premise or prior deduction. The definitions of these core concepts are presented in Section~\ref{sec:core_def}.

\eat{Let $x$ denote a logic problem instance consisting of (i) a set of natural-language premises or clues $\mathcal{C}$, and (ii) a set of domain constraints $\Gamma_{\mathrm{dom}}$ encoding the structural rules of the problem, such as exclusivity, uniqueness, admissible assignments, or variable ranges. We define the \emph{problem theory} as
\[
\Gamma_{\mathrm{base}} = \Gamma_{\mathrm{dom}} \cup \mathcal{C}.
\]

Given an input problem $x$, the goal is to train a language model policy $\pi_{\theta}$ to generate a reasoning trajectory
\[
\tau = (s_1, s_2, \dots, s_T),
\]
followed by a final solution $y$, where each step $s_t$ is \emph{interleaved} and may contain both natural-language explanation and a formal logical statement executable in Z3. Concretely, each step is represented as
\[
s_t = \bigl(u_t, z_t\bigr),
\]
where $u_t$ is a free-form natural-language rationale and $z_t$ is an optional formal statement in a constrained logical language. The full model output is therefore
\[
o = (\tau, y) \sim \pi_{\theta}(\cdot \mid x).
\]

Our objective is not only to predict the correct final solution, but to learn a policy that produces \emph{reasoning trajectories} whose intermediate formal steps are both \emph{valid} and \emph{novel}. 
Intuitively, a valid step is one that is logically supported by the underlying problem theory, while a novel step contributes new information beyond previously generated non-contradictory steps and does not merely restate an existing premise or prior deduction.

The definitions of the core concepts are presented in Section~\ref{sec:core_def}.
}

\vspace{-0.3ex}
\subsection{Core Definitions}
\label{sec:core_def}
\vspace{-0.3ex}
We first formalize the reasoning-step properties used by our solver-guided reward design. Let \(\Gamma_{\mathrm{dom}}\) denote the domain constraints of a logic problem, \(\mathcal{C}\) the set of premises, and \(\Gamma_{\mathrm{base}} = \Gamma_{\mathrm{dom}} \cup \mathcal{C}\) the resulting base theory. At reasoning step \(t\), let \(\mathcal{P}_{t-1}^{\mathrm{nc}}\) denote the set of previously accepted non-contradictory formal steps up to step \(t-1\).

\begin{definition}[Strong Implication]
\label{def:implication}
Let \(\Phi\) be a background theory and let \(A\) and \(B\) be formulas. We say that \(A\) \emph{strongly implies} \(B\) under \(\Phi\), denoted \(A \Rightarrow_{\Phi} B\), if \(\Phi \cup \{A,B\}\) is satisfiable and \(\Phi \cup \{A,\neg B\}\) is unsatisfiable. Thus, \(B\) follows from \(A\) under \(\Phi\), while vacuous implication caused by inconsistency of \(\Phi \cup \{A\}\) is excluded.
\end{definition}

\begin{definition}[Valid Step]
\label{def:valid-step}
A generated formal step \(z_t\) is \emph{valid} if it is logically implied by the base theory and remains consistent with it; that is, \(\Gamma_{\mathrm{base}} \cup \{z_t\}\) is satisfiable and \(\Gamma_{\mathrm{base}} \cup \{\neg z_t\}\) is unsatisfiable. Equivalently, \(z_t\) is a sound deduction from the premises and domain constraints.
\end{definition}

\begin{definition}[Tautological Step]
\label{def:tautology}
A generated formal step \(z_t\) is \emph{tautological} with respect to the problem if it is implied by the domain constraints alone, i.e., if \(\Gamma_{\mathrm{dom}} \cup \{\neg z_t\}\) is unsatisfiable. Such a step does not depend on the premises or any intermediate reasoning and therefore provides no problem-specific inferential progress.
\end{definition}

\begin{definition}[Novel Step]
\label{def:novel-step}
A valid step \(z_t\) is \emph{novel} if it is not tautological (Definition~\ref{def:tautology}), not already implied by the previously accepted non-contradictory reasoning steps, and not equivalent to any original premise. Let
\[
P_{t-1} \;=\; \bigwedge_{z \in \mathcal{P}_{t-1}^{\mathrm{nc}}} z
\]
denote the conjunction of all previously accepted non-contradictory formal reasoning steps, i.e., the accumulated reasoning state up to step \(t-1\). Then \(z_t\) is novel if \(P_{t-1} \not\Rightarrow_{\Gamma_{\mathrm{dom}}} z_t\) and \(z_t \not\equiv C_i\) for all \(C_i \in \mathcal{C}\), after tautological steps have been excluded according to Definition~\ref{def:tautology}. Equivalently, novelty requires that both \(\Gamma_{\mathrm{dom}} \cup \mathcal{P}_{t-1}^{\mathrm{nc}} \cup \{z_t\}\) and \(\Gamma_{\mathrm{dom}} \cup \mathcal{P}_{t-1}^{\mathrm{nc}} \cup \{\neg z_t\}\) remain satisfiable.
\end{definition}

\begin{definition}[Contradictory Step]
\label{def:contradict-step}
A generated formal step \(z_t\) is \emph{contradictory} if adding it to the current reasoning state makes the theory unsatisfiable. Equivalently, if we define the current context as \(\Gamma_{\mathrm{ctx}}^{(t)} = \Gamma_{\mathrm{base}} \cup \mathcal{P}_{t-1}^{\mathrm{nc}}\), then \(z_t\) is contradictory when
\[
\Gamma_{\mathrm{ctx}}^{(t)} \cup \{z_t\}
\text{ is unsatisfiable}.
\]
Thus, \(z_t\) conflicts with the premises, the domain constraints, or the previously accepted non-contradictory deductions.
\end{definition}

\begin{definition}[Consistency with Final Solution]
\label{def:consistent}
Let \(F\) denote the final predicted solution and let \(\mathcal{Z}\) denote the set of formal reasoning steps extracted from a trajectory. We say that \(\mathcal{Z}\) is \emph{consistent with} \(F\) if \(\Gamma_{\mathrm{dom}} \cup \mathcal{Z} \cup \{F\}\) is satisfiable. Otherwise, the reasoning trajectory is inconsistent with the final solution.
\end{definition}

\begin{figure*}[t!]
    \centering
    \includegraphics[width=1.0\textwidth]{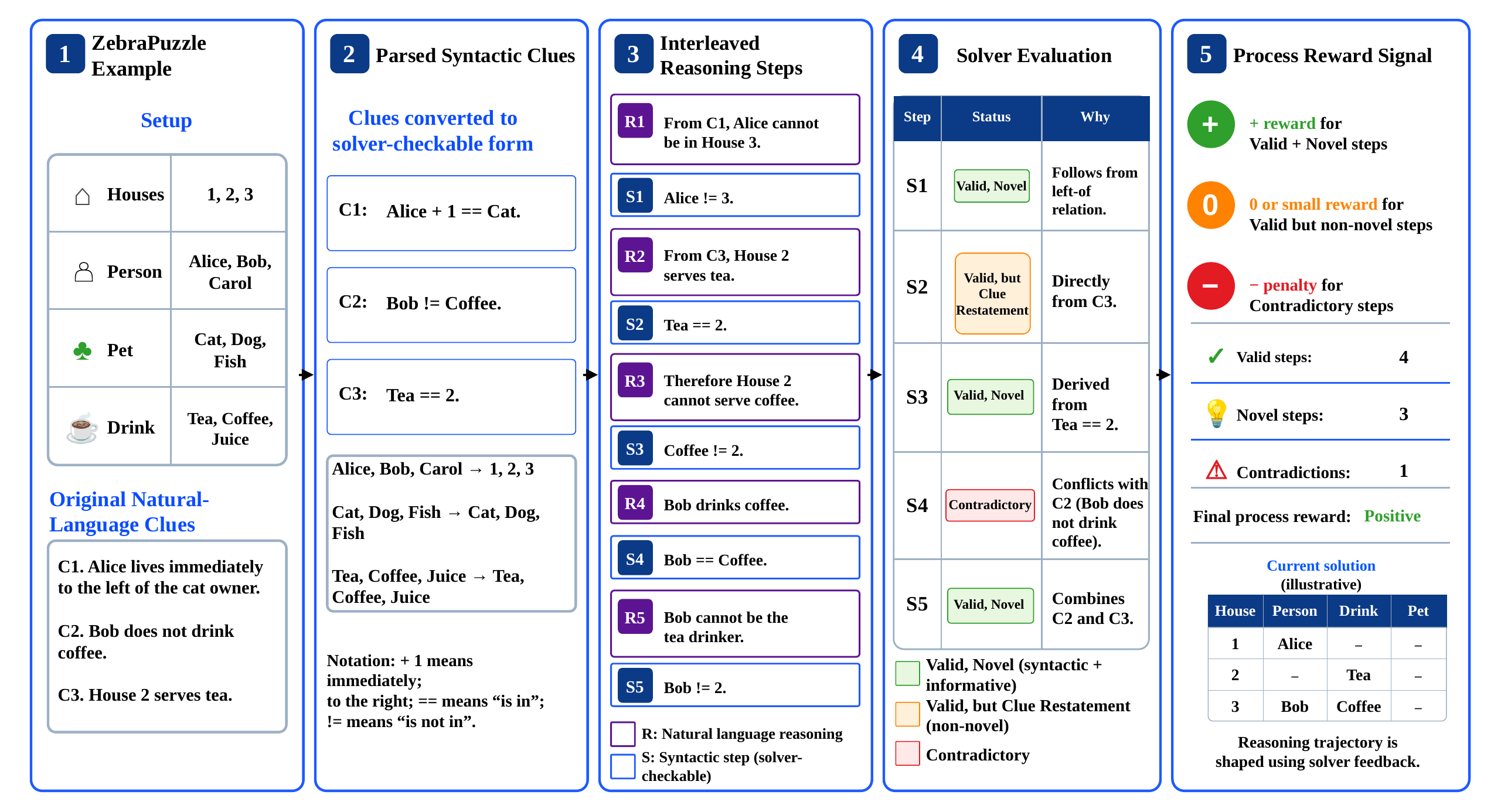}
    \vspace{-3.7ex}
    \caption{Illustration of~\OurMODEL{} using a ZebraPuzzle example. Natural-language clues are converted into solver-checkable syntactic clues, interleaved reasoning steps are classified by the solver, and the resulting signals are mapped to process rewards for training.}
    \label{fig:spring_zebra_overview}
    \vspace{-3.1ex}
\end{figure*}

\vspace{-0.7ex}
\section{\OurMODEL{}}
\label{sec:our_model}
\vspace{-0.7ex}

\noindent{\bf Overview.} The workflow of~\OurMODEL{} is illustrated in Figure~\ref{fig:spring_zebra_overview}. It is a reinforcement learning framework for improving logical reasoning in language models through symbolic verification.  It targets structured logic problems, with ZebraPuzzles used as a representative case for visualization. The workflow proceeds as follows. Given a problem instance, the LLM generates: parsed syntactic premises, interleaved reasoning steps (natural-language and formal), and a final solution. 
The parsed syntactic premises are first used to construct the symbolic solver state, after which we check whether the resulting constraint system is satisfiable.
If the solver state is SAT, it is then used to evaluate the generated formal reasoning steps and determine whether they are valid, novel, tautological, or contradictory. These solver-based judgments are converted into process rewards, encouraging reasoning trajectories that make genuine inferential progress while discouraging inconsistent, redundant, or otherwise uninformative steps.

\subsection{LLM Prompting and Output Structure}
\label{subsec:prompting}
\vspace{-0.3ex}
Given a logic problem instance \(x\), we prompt the LLM to produce a structured output that serves as the interface between free-form language reasoning and symbolic verification. Specifically, the output contains three main components: (i) \emph{Parsed Syntactic Premises}, which normalize the original natural-language premises into solver-checkable constraints, (ii) an \emph{Interleaved Reasoning trajectory}, which alternates between natural-language explanations and formal deduction steps, and (iii) a \emph{Final Solution}, which specifies the predicted assignment or answer. The prompts used in our experiments are given in Appendix~\ref{app:prompts}.

Formally, let \(\pi_{\theta}\) denote the policy LLM parameterized by \(\theta\). Given an input problem instance \(x\), the model generates a structured output:
\[
o = \bigl(\hat{\mathcal{C}}, \tau, y\bigr) \sim \pi_{\theta}(\cdot \mid x),
\]
where \(\hat{\mathcal{C}} = (\hat C_1,\ldots,\hat C_m)\) denotes the \emph{Parsed Syntactic Premises}, \(\tau=(s_1,\ldots,s_T)\) is the reasoning trajectory with \(s_t = (u_t, z_t)\), and \(y\) is the final predicted solution.\eat{Each reasoning step \(s_t\) is interleaved and represented as \(s_t = (u_t, z_t)\), where \(u_t\) is a natural-language rationale and \(z_t\) is a formal statement in a constrained symbolic language executable by the solver.} We adopt an interleaved reasoning format because LLMs are better at natural-language deduction, while formal steps are better suited for precise symbolic verification. In this setting, \(u_t\) provides the explanation and \(z_t\) the solver-checkable deduction.
\vspace{-0.3ex}
\subsection{Solver Construction and Step Verification}
\label{subsec:solver_construction}
\vspace{-0.3ex}
Given the parsed syntactic premises \(\hat{\mathcal{C}}\) generated by the policy LLM, we construct the symbolic solver by combining them with the domain constraints \(\Gamma_{\mathrm{dom}}\), yielding the base theory:
\[
\Gamma_{\mathrm{base}} = \Gamma_{\mathrm{dom}} \cup \hat{\mathcal{C}}.
\]
\eat{Here, \(\Gamma_{\mathrm{dom}}\) encodes the structural rules of the problem, such as exclusivity, uniqueness, admissible assignments, and variable ranges, while \(\hat{\mathcal{C}}\) provides the parsed premise-specific constraints. }
We first check whether \(\Gamma_{\mathrm{base}}\) is satisfiable and whether it admits a unique solution. 
For ZebraLogic, we check that the constraints admit a unique
satisfying assignment. For AR-LSAT, we instead check that
exactly one answer option is correct, while allowing the
underlying constraints to admit multiple satisfying assignments.
These checks establish that the parsed constraints are solver-compatible and yield a unique answer under the generated formalization.
If \(\Gamma_{\mathrm{base}}\) is unsatisfiable or fails to yield a unique solution, the trajectory is treated as solver-incompatible and receives a reduced reward.

\paragraph{Symbolic Step Verification.}
Once the base solver state is established, we use it to verify each parsed formal reasoning step \(z_t\). Let \(\mathcal{P}_{t-1}^{\mathrm{nc}}\) denote the set of previously accepted non-contradictory steps, and define the current reasoning context as
\[
\Gamma_{\mathrm{ctx}}^{(t)} = \Gamma_{\mathrm{base}} \cup \mathcal{P}_{t-1}^{\mathrm{nc}}.
\]
For each generated step \(z_t\), we extract the core solver signals that drive our reward design. First, we check whether \(z_t\) is \emph{valid} according to Definition~\ref{def:valid-step}, i.e., whether it is entailed by \(\Gamma_{\mathrm{base}}\) while remaining satisfiable with it. Second, we check whether \(z_t\) is \emph{tautological} according to Definition~\ref{def:tautology}, namely whether it follows from \(\Gamma_{\mathrm{dom}}\) alone and therefore contributes no problem-specific information. Third, for a valid and non-tautological step, we test whether it is \emph{novel} using Definition~\ref{def:novel-step}. Concretely, letting \(P_{t-1}=\bigwedge_{z\in\mathcal{P}_{t-1}^{\mathrm{nc}}} z\), we check whether
\[
P_{t-1} \not\Rightarrow_{\Gamma_{\mathrm{dom}}} z_t,
\]
and whether \(z_t\) is not equivalent to any original premise \(C_i \in \mathcal{C}\). This step-level novelty signal is aggregated into the reward term \(r_{\text{novel}}\), which measures the extent to which the trajectory introduces new inferential content beyond the previously accepted reasoning state. Fourth, we check whether \(z_t\) is \emph{contradictory} according to Definition~\ref{def:contradict-step}, i.e., whether
\[
\Gamma_{\mathrm{ctx}}^{(t)} \cup \{z_t\}
\]
is unsatisfiable. The corresponding contradiction signal is accumulated into \(r_{\text{contra}}\), which penalizes trajectories that introduce logically inconsistent deductions. Finally, at the trajectory level, we check whether the generated reasoning remains \emph{consistent with the final solution} using Definition~\ref{def:consistent}, namely whether the formal reasoning steps and the final prediction can coexist in a satisfiable theory. This trajectory-level compatibility signal is mapped to \(r_{\text{consistency}}\), which rewards reasoning traces whose intermediate deductions remain coherent with the final predicted solution.

These solver-derived signals, namely novelty, contradiction, and final-solution consistency, provide the fine-grained process supervision used by~\OurMODEL{}. Intuitively, they allow the model to distinguish between steps that make genuine inferential progress and those that are redundant, uninformative, or logically inconsistent.

\subsection{Process-Level Reward}
\label{sec:process_reward}

We define a structured reward that combines solver-derived process signals (Section~\ref{subsec:solver_construction}) with task-level and formatting signals. Concretely, our reward uses the core symbolic signals of \emph{novelty}, \emph{contradiction}, and \emph{final-solution consistency}, together with three auxiliary signals:
(i) \(r_{\text{format}}\), a binary reward that takes value \(1\) if the generated reasoning follows the required interleaved structure of natural-language and parsed formal steps, and \(0\) otherwise;
(ii) \(r_{\text{parsed}}\), a binary reward that takes value \(1\) if the LLM-generated output can be successfully parsed into the target symbolic representation, and \(0\) otherwise; and (iii) \(r_{\text{accuracy}}\), which measures final accuracy. \textbf{Note}, for ZebraPuzzles, we instantiate \(r_{\text{accuracy}}\) as puzzle accuracy.
We distinguish two cases using the satisfiability indicator \(s_{\text{sat}} \in \{0,1\}\) of the parsed premises system. If \(s_{\text{sat}}=0\), the reward is restricted to auxiliary quality signals and final-answer quality; if \(s_{\text{sat}}=1\), we additionally apply process-level rewards from symbolic step verification:
\[
    R =
    \begin{cases}
        r_{\mathrm{base}}, & s_{\mathrm{sat}}=0,\\
        r_{\mathrm{base}}
        +(0.5+0.5r_{\mathrm{accuracy}})r_{\mathrm{proc}},
        & s_{\mathrm{sat}}=1,
    \end{cases}
\]
where
\[
    r_{\mathrm{base}}
    =0.15r_{\mathrm{parsed}}
    +0.10r_{\mathrm{format}}
    +0.60r_{\mathrm{accuracy}},
\]
\[
    r_{\mathrm{proc}}
    =0.40r_{\mathrm{novel}}
    +0.30r_{\mathrm{consistency}}
    -0.15r_{\mathrm{contra}}.
\]
This reward design preserves the same base accuracy reward regardless of $s_{\mathrm{sat}}$, while adding process supervision based on solver-verified reasoning quality when available.
As a result, trajectories with more novel and solution-consistent deductions, and fewer contradictions, receive higher reward, aligning reinforcement learning with both \emph{outcome quality} and \emph{reasoning quality}.
\eat{
This reward design serves two purposes. First, it preserves the main task objective through \(r_{\text{accuracy}}\), while also encouraging outputs that are structurally usable by the symbolic verifier through \(r_{\text{format}}\) and \(r_{\text{parsed}}\). Second, once the solver state is satisfiable, it introduces process supervision through solver-verified reasoning quality, so that trajectories with more novel and solution-consistent deductions, and fewer contradictions, receive higher reward. In this way, \OurMODEL{} aligns reinforcement learning with both \emph{outcome quality} and \emph{reasoning quality}.
}

\vspace{-0.3ex}
\subsection{RL Training}
\label{subsec:RL-training}
\vspace{-0.3ex}
We train~\OurMODEL{} with a group-based reinforcement learning objective, using GRPO for policy optimization and a DAPO-style distributed runtime for scalable sampling and training~\citep{shao2024deepseekmath,yu2025dapo}. Given a logic problem \(x\), the policy \(\pi_{\theta}\) samples a group of candidate outputs \(\{o_i\}_{i=1}^{G}\), where each \(o_i=(\tau_i,y_i)\) contains an interleaved reasoning trajectory \(\tau_i\) and a final solution \(y_i\). Each candidate receives a scalar reward \(r_i = R(x,o_i)\) from our solver-guided verifier. The training objective is to maximize the expected reward:
\begin{equation}
\max_{\theta}\; \mathbb{E}_{x \sim \mathcal{D},\, o \sim \pi_{\theta}(\cdot \mid x)}\bigl[R(x,o)\bigr].
\end{equation}
Following GRPO, rewards are normalized within each sampled group. Let \(\bar r=\frac{1}{G}\sum_{i=1}^{G} r_i\) and \(\sigma_r=\sqrt{\frac{1}{G}\sum_{i=1}^{G}(r_i-\bar r)^2}\). The group-relative advantage is then \(A_i=\frac{r_i-\bar r}{\sigma_r+\epsilon}\), where \(\epsilon>0\) ensures numerical stability. Using the policy ratio \(\rho_i(\theta)=\frac{\pi_{\theta}(o_i\mid x)}{\pi_{\theta_{\mathrm{old}}}(o_i\mid x)}\), the clipped GRPO loss is:
\begin{equation}
\mathcal{L}_{\mathrm{GRPO}}(\theta)
=
\mathbb{E}_{x}\!\left[
\frac{1}{G}\sum_{i=1}^{G}
\min\!\bigl(
\rho_i(\theta)A_i,\,
\operatorname{clip}(\rho_i(\theta),1-\delta,1+\delta)A_i
\bigr)
\right]
\end{equation}
This encourages the policy to prefer candidates with better solver-guided reasoning quality relative to others in the same group. To stabilize training, we regularize the policy toward a frozen reference policy \(\pi_{\mathrm{ref}}\):
\begin{equation}
\mathcal{L}(\theta)
=
\mathcal{L}_{\mathrm{GRPO}}(\theta)
-
\beta\,\mathrm{KL}\!\bigl(\pi_{\theta}(\cdot\mid x)\,\|\,\pi_{\mathrm{ref}}(\cdot\mid x)\bigr),
\end{equation}
where \(\beta\) controls the regularization strength. In this framework, DAPO provides the distributed runtime for large-scale sampling, reward computation, and optimization, while Z3 solver~\citep{de2008z3} supplies fine-grained process feedback over intermediate deductions. 

\color{black}

\vspace{-0.7ex}
\section{Experimentation}
\label{sec:experiments}
\vspace{-0.7ex}
\subsection{Experimental Settings}
\noindent{\bf (a) Datasets.}

For evaluation, we use three logical reasoning benchmarks. 
(i) \textbf{ZebraLogic}~\citep{lin2025zebralogic} contains 1,000 Zebra puzzles across four difficulty levels: Small, Medium, Large, and Extra-Large. We use a 25\%/5\%/70\% train/validation/test split, corresponding to 250/50/700 instances.
(ii) \textbf{AR-LSAT}~\citep{zhong2022analytical} consists of analytical reasoning problems across three categories: Ordering, Grouping, and Assignment. We use 300 training and 50 validation instances per category. For testing, we use the official 230-instance test split, comprising 112 Ordering, 49 Grouping, and 69 Assignment problems.
(iii) \textbf{Knights and Knaves (KnK)}~\citep{xie2024memorization} contains 6,200 training and 700 test puzzles spanning 2--8 characters. We sample 300 puzzles from the training set for RL model training and hold out an additional 50 for validation. We evaluate on the complete 700-puzzle test set. Dataset statistics are summarized in Table~\ref{tab:dataset_stats}.

\noindent{\bf (b) Large Models.}
We evaluate~\OurMODEL{} on Qwen3-1.7B (non-thinking), Qwen3-4B-Thinking, and Qwen3-8B~\citep{yang2025qwen3}, as well as Phi-4-Reasoning~\citep{abdin2025phi4reasoning}. The first two cover contrasting small-model configurations (i.e., thinking and non-thinking), while the latter two extend the evaluation to a larger model and a different model family.

\noindent{\bf (c) Baselines.} We consider the following baselines. \underline{(i) \textbf{Base-NL}} and \underline{\textbf{Base-INT}} denote the pretrained LLMs without task-specific reinforcement learning, using a natural-language prompt and an interleaved prompt, respectively. To isolate the contribution of solver-guided process supervision, we additionally compare against two outcome-only reward baselines. \underline{(ii) \textbf{OR-NL}} uses the same natural-language prompt as Base-NL, but optimizes only for final-answer quality with reward \(0.15 \cdot r_{\text{parsed}} + 0.6 \cdot r_{\text{accuracy}}\). \underline{(iii) \textbf{OR-INT}} uses the same interleaved prompt as Base-INT, but likewise optimizes only for final-answer quality with reward \(0.15 \cdot r_{\text{parsed}} + 0.6 \cdot r_{\text{accuracy}}\). Thus, OR-NL and OR-INT remove the process-level reward terms of~\OurMODEL{} and differ only in the prompting format. From existing research on logical reasoning with LLMs, we use Logic-LM~\citep{pan2023logiclm} as a representative baseline for comparison.

\noindent{\bf (d) Experimental Settings.}
For RL training, we use the following settings. The GRPO group size $n$ is set to \(8\), the sampling temperature during training is \(0.8\), and the inference temperature is fixed at \(0.0\). We set \(\beta = 0.01\), and the maximum input and output lengths to \(8 \times 1024\) tokens. All experiments are conducted using the VERL framework\footnote{\url{https://verl.readthedocs.io/}} on a cluster with \(4\times\) NVIDIA H200 GPUs. We use the Z3 solver for symbolic-step verification~\citep{de2008z3}. Under this configuration, training Qwen3-4B-Thinking requires approximately 45 minutes per epoch. Unless stated otherwise, training is run for up to 25 epochs with early stopping using a patience of 3 epochs, and we report the checkpoint with the best validation performance. All experiments are repeated over 3 runs, and we report average scores. The prompt templates used in our experiments are provided in Appendix~\ref{app:prompts}.

\noindent{\bf (e) Evaluation Metrics.}
We evaluate model performance using task-specific metrics. For ZebraLogic, we report \textbf{Puzzle Acc.} and \textbf{Cell Acc.}; for AR-LSAT, we report \textbf{Acc.}; and for Knights and Knaves, we report \textbf{Puzzle Acc.} and \textbf{Person Acc.}. Further details and the mathematical definitions of these metrics are provided in Appendix~\ref{app:evaluation-metrics}.

\begin{table*}[t]
\centering
\caption{\OurMODEL{} performance across different LLMs on ZebraLogic, AR-LSAT, and Knights and Knaves. We report standard deviation for \OurMODEL{} and its ablation. AR-LSAT
Overall accuracy is the weighted mean of Ordering,
Grouping, and Assignment accuracies.}
\label{tab:combined_results}
\small
\setlength{\tabcolsep}{3.5pt}
\resizebox{\textwidth}{!}{%
\begin{tabular}{llcc|cccc|cc}
\toprule
\multirow{2}{*}{\textbf{LLM}}
& \multirow{2}{*}{\textbf{System}}
& \multicolumn{2}{c|}{\textbf{ZebraLogic}}
& \multicolumn{4}{c|}{\textbf{AR-LSAT (Acc.)}}
& \multicolumn{2}{c}{\textbf{Knights and Knaves}} \\
\cmidrule(lr){3-4}
\cmidrule(lr){5-8}
\cmidrule(lr){9-10}
&
& \textbf{Puzzle Acc.} $\uparrow$
& \textbf{Cell Acc.} $\uparrow$
& \textbf{Order.} $\uparrow$
& \textbf{Group.} $\uparrow$
& \textbf{Assign.} $\uparrow$
& \textbf{Overall} $\uparrow$
& \textbf{Puzzle Acc.} $\uparrow$
& \textbf{Person Acc.} $\uparrow$ \\
\midrule

\multirow{7}{*}{Qwen3-1.7B}
& Base-NL & 11.14 & 35.04  & 16.07 & 24.48 & 0.00 & 13.04 & 53.28 & 64.43 \\
& OR-NL   & 38.57 & 49.87  & 19.64 & 40.81 & 1.44 & 18.69 & 68.57 & 79.64 \\
& Base-INT & 31.85 & 41.45 & 23.21 & 24.48 & 14.49 & 20.86 & 18.71 & 55.61 \\
& OR-INT   & 36.00 & 46.09 & 35.71 & 51.02 & 24.63 & 35.65 & 73.57 & 80.54 \\
\cmidrule(lr){2-10}
& Logic-LM    & 12.85 & 33.16 & - & - & - & 15.58 & 3.57 & 22.51 \\
\cmidrule(lr){2-10}
& \OurMODEL{} (--N) & 19.28 (±0.92) & 59.62 (±0.81) & 34.82 & 55.10 & 46.37 & 42.61 (±0.95) & 78.57 (±0.97) & 82.14 (±0.95) \\
& \textbf{\OurMODEL{}}
& \textbf{54.00 (±0.62)} & \textbf{59.87 (±0.65)}
& \textbf{41.07} & \textbf{57.14} & \textbf{49.56} & \textbf{47.04 (±0.66)}
& \textbf{85.28 (±0.90)} & \textbf{86.94 (±0.87)} \\
\midrule

\multirow{7}{*}{Qwen3-4B-Thinking}
& Base-NL  & 24.00 & 24.00 & 6.25  & 18.36 & 0.00  & 6.95  & 55.28 & 65.41 \\
& OR-NL    & 67.28 & 68.30 & 25.00 & 65.30 & 10.14 & 29.13 & 74.28 & 84.66 \\
& Base-INT & 17.00 & 23.32 & 6.26  & 14.28 & 5.50  & 7.74  & 33.14 & 63.47 \\
& OR-INT   & 66.14 & 72.72 & 61.60 & 79.59 & 65.21 & 66.52 & 76.57 & 87.58 \\
\cmidrule(lr){2-10}
& Logic-LM    & 21.42 & 38.12 & - & - & - & 20.78 & 7.14 & 35.48 \\
\cmidrule(lr){2-10}
& \OurMODEL{} (--N) & 61.42 (±0.84) & 68.27 (±0.73) & 63.39 & 77.55 & 66.66 & 67.39 (±0.87) & 82.85 (±0.88)& 90.13 (±0.85) \\
& \textbf{\OurMODEL{}}& \textbf{73.71 (±0.51)} & \textbf{81.52 (±0.53)} & \textbf{69.64} & \textbf{81.63} & \textbf{69.56} & \textbf{72.17 (±0.76)} & \textbf{88.71 (±0.85)} & \textbf{94.21 (±0.79)} \\
\midrule

\multirow{7}{*}{Phi4-Reasoning}
& Base-NL
            & 7.14  & 22.42 & 12.50 & 24.48 & 17.39 & 16.52 & 57.85 & 68.13 \\
& OR-NL     & 47.14 & 51.27 & 21.42 & 40.81 & 30.43 & 28.25 & 72.85 & 85.57 \\
& Base-INT  & 18.71 & 20.27 & 0.00 & 18.36 & 2.89 & 4.78 & 36.42 & 68.84 \\
& OR-INT    & 49.00 & 53.54 & 23.21 & 46.93 & 26.08 & 29.12 & 81.42 & 88.13 \\
\cmidrule(lr){2-10}
& Logic-LM    & 20.0 & 36.75 & - & - & - & 18.61 & 9.28 & 40.12 \\
\cmidrule(lr){2-10}
& \OurMODEL{} (--N) & 53.71 (±1.15) & 56.34 (±1.17) & 22.32 & 42.85 & 24.63 & 27.39 (±1.04) & 83.57 (±1.10) & 88.51 (±0.95) \\
& \textbf{\OurMODEL{}}
& \textbf{55.71 (±1.10)} & \textbf{59.52 (±1.07)}
& \textbf{26.78} & \textbf{51.02}
& \textbf{34.78} & \textbf{34.34 (±1.03)}
& \textbf{89.28 (±0.99)} & \textbf{94.57 (±0.92)} \\
\midrule

\multirow{7}{*}{Qwen3-8B}
& Base-NL   & 47.42 & 50.13 & 29.46 & 40.81 & 30.43 & 32.17 & 56.0 & 68.49 \\
& OR-NL     & 65.67 & 64.85 & 51.78 & 61.22 & 56.52 & 55.21 & 74.28 & 80.76 \\
& Base-INT  & 50.42 & 58.61 & 33.92 & 30.61 & 34.78 & 33.47 & 49.57 & 72.14 \\
& OR-INT    & 70.85 & 75.13 & 65.17 & 77.08 & 63.76 & 67.28 & 77.14 & 82.85 \\
\cmidrule(lr){2-10}
& Logic-LM  & 22.85 & 40.56 & - & - & - & 25.97 & 8.57 & 55.20 \\
\cmidrule(lr){2-10}
& \OurMODEL{} (--N) & 69.42 (±0.80) & 76.57 (±0.79) & 51.78 & 75.51 & 50.72 & 56.52 (±0.72) & 87.14 (±0.84) & 91.24 (±0.81) \\
& \textbf{\OurMODEL{}} & \textbf{77.85 (±0.78)} & \textbf{79.41 (±0.74)} & \textbf{71.42} & \textbf{85.71} & \textbf{69.56} & \textbf{73.91 (±0.69)} & \textbf{93.14 (±0.95)} & \textbf{96.05 (±0.90)} \\
\bottomrule
\end{tabular}%
}
\vspace{-3.7ex}
\end{table*}
\vspace{-0.3ex}
\subsection{Main Results}
\label{sec:results}
\vspace{-0.3ex}

The results in Table~\ref{tab:combined_results} show that \OurMODEL{} consistently improves logical reasoning across four LLMs and three benchmarks. The gains hold across models of different sizes and reasoning capabilities, including Qwen3-1.7B, Qwen3-4B-Thinking, Phi4-Reasoning, and Qwen3-8B. In contrast, the base models exhibit substantial sensitivity to prompting style, while outcome-only reward learning improves performance but generally remains below \OurMODEL{}. Logic-LM also performs substantially below \OurMODEL{} across all evaluated model--benchmark combinations.

\noindent{\bf (a) ZebraLogic.}
\OurMODEL{} achieves strong gains across all four LLMs, reaching Puzzle Acc scores of 54.00, 73.71, 55.71, and 77.85 for Qwen3-1.7B, Qwen3-4B-Thinking, Phi4-Reasoning, and Qwen3-8B, respectively. For Qwen3-1.7B, \OurMODEL{} improves Puzzle Acc from 38.57 for the strongest outcome-only baseline (OR-NL) to 54.00, a gain of 15.43 points. For Qwen3-4B-Thinking, it reaches 73.71 Puzzle Acc and 81.52 Cell Acc, compared with 67.28 and 72.72 for the strongest outcome-only results on the respective metrics. The improvements remain evident for other LLMs with \OurMODEL{} reaching 55.71/59.52 Puzzle/Cell Acc with Phi4-Reasoning and 77.85/79.41 with Qwen3-8B. Logic-LM performs considerably worse, with Puzzle Acc ranging from 12.85 to 22.85 across the four LLMs. These results indicate that solver-guided process optimization provides benefits beyond prompting, outcome-only optimization, and existing solver-augmented logical reasoning.

The novelty reward is particularly important on ZebraLogic. Removing it (\OurMODEL{} (--N)) reduces Puzzle Acc from 54.00 to 19.28 for Qwen3-1.7B, from 73.71 to 61.42 for Qwen3-4B-Thinking, from 55.71 to 53.71 for Phi4-Reasoning, and from 77.85 to 69.42 for Qwen3-8B. The effect varies across models, but the consistent reduction in puzzle-level accuracy shows that rewarding novel deductions complements solver-validity and helps translate locally valid reasoning into complete puzzle solutions.

\noindent{\bf (b) AR-LSAT.}
\OurMODEL{} also consistently improves performance on AR-LSAT, achieving overall accuracies of 47.04, 72.17, 34.34, and 73.91 for Qwen3-1.7B, Qwen3-4B-Thinking, Phi4-Reasoning, and Qwen3-8B, respectively. For Qwen3-1.7B, the strongest outcome-only baseline, OR-INT, achieves 35.65 overall accuracy, whereas \OurMODEL{} reaches 47.04, with the largest improvement on Assignment (24.63 to 49.56). For Qwen3-4B-Thinking, \OurMODEL{} improves the overall score from 66.52 for OR-INT to 72.17, while for Qwen3-8B it improves from 67.28 to 73.91. Phi4-Reasoning exhibits lower absolute AR-LSAT performance, but \OurMODEL{} still improves over its strongest outcome-only result from 29.12 to 34.34. Removing the novelty reward reduces overall accuracy for all four LLMs, further supporting the contribution of novel-step supervision beyond solver-validity alone.

\noindent{\bf (c) Knights and Knaves.}
The same trend extends to Knights and Knaves, where \OurMODEL{} achieves the strongest Puzzle Acc and Person Acc for all four LLMs. Puzzle Acc reaches 85.28, 88.71, 89.28, and 93.14 for Qwen3-1.7B, Qwen3-4B-Thinking, Phi4-Reasoning, and Qwen3-8B, respectively, with corresponding Person Acc scores of 86.94, 94.21, 94.57, and 96.05. Compared with the strongest outcome-only baseline for each model, this corresponds to Puzzle Acc gains of 11.71, 12.14, 7.86, and 16.00 points, respectively. Logic-LM performs particularly poorly at puzzle-level exact solving, achieving only 3.57--9.28 Puzzle Acc. Moreover, removing the novelty reward consistently reduces Puzzle Acc, from 85.28 to 78.57, 88.71 to 82.85, 89.28 to 83.57, and 93.14 to 87.14 across the four models. Together, these results demonstrate that the benefits of solver-guided process rewards and novelty supervision generalize across model families, model scales, and structurally different logical reasoning tasks.

Across three runs, \OurMODEL{} exhibits low run-to-run variability,
with standard deviations ranging from 0.51 to 1.10 percentage
points. Even the largest reported standard deviation is small
relative to the gains over the strongest outcome-only baselines,
supporting the consistency of the observed improvements.

\vspace{-0.3ex}
\subsection{Ablation Analyses}
\label{sec:ablation}
\vspace{-0.3ex}
To assess the contribution of the key reward components in~\OurMODEL{}, we conduct an ablation study for different model components on ZebraLogic benchmark. Specifically, 
(i) \texttt{--N} denotes the variant without the novel-step reward; 
(ii) \texttt{--NC} removes both the novel-step reward and the contradiction penalty; 
(iii)~\OurMODEL{}, i.e., the full model.
We compare these variants to quantify how each component contributes to improved reasoning trajectories and final performance.

Table~\ref{tab:combined_trace_quality} compares the three reward settings from both an outcome-level and a process-level perspective. \textbf{Puzzle Acc.} measures end-task success. The columns under \textbf{Solved traces} summarize the quality of reasoning trajectories on correctly solved cases only, including \textbf{SAT-rate}, which measures solver consistency of successful traces, \textbf{Steps}, which measures reasoning efficiency, and \textcolor{black}{\textbf{Parsability} (Parsed symbolic steps / Total symbolic steps)} and \textcolor{black}{\textbf{Validity rate} (Valid symbolic steps / Parsed symbolic steps)}, which quantify structural well-formedness and logical correctness of the generated steps. The columns under \textbf{Failed traces} characterize the behavior of unsuccessful trajectories, including how long the model continues reasoning after failure, how structurally usable those traces remain, and how often they introduce contradictions. Taken together, these fields allow us to evaluate not only whether a system solves the puzzle, but also how efficient, reliable, and solver-faithful its reasoning process is.

\begin{table*}[t]
\centering
\caption{Reasoning-trace quality across three reward settings for Qwen3-4B-Thinking. \textbf{Solved traces} are averaged over correctly solved cases, while \textbf{Failed traces} are averaged over incorrect cases.}
\label{tab:combined_trace_quality}
\small
\setlength{\tabcolsep}{4pt}
\resizebox{\textwidth}{!}{%
\begin{tabular}{l c c c c c c c c c}
\toprule
\multirow{2}{*}{\textbf{System}} & \multirow{2}{*}{\textbf{Puzzle Acc.} \(\uparrow\)} 
& \multicolumn{4}{c}{\textbf{Solved traces}} 
& \multicolumn{4}{c}{\textbf{Failed traces}} \\
\cmidrule(lr){3-6} \cmidrule(lr){7-10}
& 
& \textbf{SAT-rate \(\uparrow\)} 
& \textbf{Steps \(\downarrow\)} 
& \textbf{Parsability \(\uparrow\)} 
& \textbf{Validity rate \(\uparrow\)}
& \textbf{Steps \(\downarrow\)} 
& \textbf{Parsability \(\uparrow\)} 
& \textbf{Validity rate \(\uparrow\)} 
& \textbf{Contradictions / 100 Steps \(\downarrow\)} \\
\midrule
\texttt{--N}    & 61.42 & 0.94 & 35.20 & 0.92 & 0.98 & 49.10 & 0.76 & 0.82 & 6.77 \\
\texttt{--NC}   & 66.14 & 0.81 & 34.82 & 0.69 & 0.95 & 11.23 & 0.61 & 0.91 & 2.52 \\
\OurMODEL{}     & \textbf{73.71} & \textbf{0.96} & \textbf{22.37} & \textbf{0.98} & \textbf{0.99} & \textbf{9.39} & \textbf{0.99} & \textbf{0.99} & \textbf{0.46} \\
\bottomrule
\end{tabular}}
\vspace{-4.1ex}
\end{table*}

\noindent{\bf (a) Removing novel-step rewards (\texttt{--N}).}
Compared with \OurMODEL{}, the (\texttt{--N}) setting achieves substantially lower Puzzle Acc.\ (\(61.42\) vs.\ \(73.71\)) and requires much longer solved traces (35.20 vs.\ 22.37 steps). Its failed traces are especially poor: when the model is incorrect, it continues generating for an average of 49.10 steps and accumulates 6.77 contradictions per 100 steps. In addition, both its Parsability and Validity Rate are consistently worse than those of \OurMODEL{}, for solved as well as failed cases. This indicates that without the novel-step reward, reasoning becomes less efficient, less stable, and far more contradiction-prone.

\noindent{\bf (b) Removing both novelty and contradiction (\texttt{--NC}).}
The (\texttt{--NC}) setting yields lower performance compared to~\OurMODEL{}. Its solved traces have the lowest SAT-rate (\(0.81\)), the worst Parsability (\(0.69\)), and a noticeably lower Validity Rate (\(0.95\)) than both alternatives. Its failed traces are also substantially noisier than those of \OurMODEL{}, with a contradiction rate of 2.52 per 100 steps. Thus, while (\texttt{--NC}) improves over the novelty-free baseline in terms of trace length and final accuracy, it still fails to maintain solver-faithful and structurally reliable reasoning trajectories.

A possible explanation for the observed ordering between
\OurMODEL{} (\texttt{--NC}) and \OurMODEL{} (\texttt{--N})
is the asymmetric role of the reward components:
the contradiction term penalizes undesirable steps, whereas
the novelty term provides positive guidance toward useful
inferential progress. Without the novelty reward, retaining
the contradiction penalty may provide insufficient constructive
guidance during optimization, potentially explaining why
\OurMODEL{} (\texttt{--N}) underperforms
\OurMODEL{} (\texttt{--NC}) in some settings.
In the full model, novelty rewards may provide constructive
guidance, with the contradiction penalty acting as a
complementary regularizer.

The two ablations also exhibit different failure behavior.
On ZebraLogic, \OurMODEL{} (\texttt{--N}) produces longer
failed traces than \OurMODEL{} (\texttt{--NC}), averaging
49.10 versus 11.23 steps, and has a higher contradiction
rate of 6.77 versus 2.52 per 100 steps. These observations
suggest different failure patterns; the lower contradiction
rate of \OurMODEL{} (\texttt{--NC}) does not by itself
establish better reasoning. Since these rates are normalized
by trace length, longer traces alone cannot explain the
difference.

\noindent{\bf (c) Impact of novel steps (\OurMODEL).}
\OurMODEL{} is the strongest system across both final performance and trace quality. It attains the highest Puzzle Acc.\ (\(73.71\)) and the highest SAT-rate on solved cases (\(0.96\)), indicating that its successful trajectories are most compatible with solver-based verification. It also reaches correct solutions with the shortest solved traces and the best Parsability and Validity Rate, showing that its reasoning is both more concise and more structurally reliable. The advantage is equally clear on failed cases: \OurMODEL{} has the shortest failed traces (9.39 steps), the highest Parsability and Validity Rate, and by far the lowest contradiction rate (0.46 per 100 steps). Relative to both (\texttt{--N}) and (\texttt{--NC}), these results show that solver-guided process supervision does not merely improve whether the model gets the answer right; it also regularizes the entire reasoning trajectory toward shorter, cleaner, and more solver-faithful traces.

\vspace{-0.3ex}
\subsection{Further Analyses}
\label{sec:further_analyses}
\vspace{-0.3ex}
To better understand the behavior of~\OurMODEL{}, we perform a set of additional analyses on the ZebraLogic benchmark, including: 
\textbf{(a)} Scaling Behavior of~\OurMODEL{} across different puzzle sizes;
\textbf{(b)} Solver SAT rate across Epochs;
\textbf{(c)} Error Analyses; and 
\textbf{(d)} Qualitative case study. 
Corresponding results, together with the core findings, are reported in Appendix~\ref{app:exp_addl}. These analyses complement the main results by revealing how~\OurMODEL{} behaves under increasing puzzle difficulty, how its solver consistency develops during training, and how its reasoning traces remain qualitatively cleaner and more stable than those of the baselines.

\vspace{-0.7ex}
\section{Conclusion and Future Work}
\label{sec:conclusion}
\vspace{-0.7ex}
We introduced~\OurMODEL{}, a solver-guided reinforcement learning framework
that provides process-level supervision by rewarding novel, logically
consistent deductions and penalizing contradictory and uninformative steps.
Across three benchmarks---ZebraLogic, AR-LSAT, and Knights and Knaves---and
four LLMs, \OurMODEL{} consistently outperforms base LLMs, outcome-only
reward baselines, and Logic-LM. Our analyses show shorter successful
reasoning traces, fewer contradictions, and greater consistency with
logical constraints. Future work will extend \OurMODEL{} to broader
reasoning domains, including code generation, using suitable formalizations
and verification procedures for intermediate steps.

\section*{Acknowledgments}
The authors gratefully acknowledge the computing support and resources provided by the Panther high-performance computing cluster at the Qatar Computing Research Institute (QCRI).

\bibliographystyle{plainnat}
\bibliography{our_refer}

\clearpage
\appendix

\section{Background}
\label{app:background}

We consider three representative logical reasoning benchmarks in this work:
ZebraLogic, AR-LSAT, and Knights and Knaves. These benchmarks require
multi-step deduction under explicit logical constraints but differ substantially
in structure and output format. ZebraLogic emphasizes exact assignment under
global consistency constraints, AR-LSAT evaluates textual reasoning over
ordering, grouping, and assignment scenarios, and Knights and Knaves requires
truth-conditional reasoning over multiple characters. Together, these
benchmarks provide complementary settings for evaluating logically valid and
informative intermediate reasoning~\citep{lin2025zebralogic,zhong2022analytical}. 

\subsection{ZebraPuzzles}
\label{app:zebrapuzzles}

ZebraPuzzles, also known as logic grid puzzles, are structured reasoning problems in which the goal is to determine a unique assignment of entities to attributes using a set of natural-language clues together with domain constraints. A typical puzzle specifies a fixed number of positions or houses and several attribute categories, such as person, pet, drink, or color. Each value in an attribute category must be assigned exactly once, and each position must receive exactly one value from each category. Solving the puzzle therefore requires satisfying both clue-derived constraints and global one-to-one assignment constraints. ZebraLogic formalizes this setting as a benchmark for studying logical reasoning under controllable complexity, with puzzles generated from underlying constraint satisfaction problems~\citep{lin2025zebralogic}. 

From the perspective of formal reasoning, ZebraPuzzles are particularly attractive because they admit a clean symbolic representation. Clues such as equality, inequality, adjacency, left--right relations, and exclusivity can be encoded as logical constraints, while the final solution corresponds to a complete satisfying assignment. This makes the benchmark especially suitable for our setting, where a symbolic solver can verify whether an intermediate deduction is valid, contradictory, or genuinely contributes new information to the evolving reasoning state. ZebraLogic further shows that performance drops sharply as puzzle complexity increases, indicating that these puzzles remain challenging even for strong modern LLMs~\citep{lin2025zebralogic}. 

\noindent{\bf Example.}
Consider a simple Zebra-style puzzle with three houses and three attribute categories:
\begin{align*}
\textit{Person} & = \{\text{Alice}, \text{Bob}, \text{Carol}\}, \\
\textit{Pet}    & = \{\text{Cat}, \text{Dog}, \text{Fish}\}, \\
\textit{Drink}  & = \{\text{Tea}, \text{Coffee}, \text{Juice}\}.
\end{align*}
Each house contains exactly one person, one pet, and one drink, and each attribute value is used exactly once.

\medskip
\noindent{\bf Clues.}
\begin{enumerate}
    \item Alice lives immediately to the left of the cat owner.
    \item Bob lives in the house where tea is served.
    \item The person in House 2 drinks tea.
    \item Carol owns the fish.
    \item The dog owner drinks coffee.
    \item Bob does not drink coffee.
\end{enumerate}

\medskip
To solve this puzzle, the model may proceed through the following intermediate reasoning steps.

\noindent{\bf Reasoning Steps.}
\begin{enumerate}
    \item From Clues 2 and 3, Bob must live in House 2.
    \item From Clue 1, Alice cannot live in House 3, since there is no house to its right.
    \item Since Bob already occupies House 2, Alice cannot live in House 2. Therefore, Alice must live in House 1, and Carol must live in House 3.
    \item By Clue 1, the cat owner must live immediately to the right of Alice. Since Alice is in House 1, the cat owner must be in House 2. Hence Bob owns the cat.
    \item From Clue 4, Carol owns the fish, so the remaining pet, the dog, must belong to Alice in House 1.
    \item From Clue 5, the dog owner drinks coffee. Therefore, Alice drinks coffee.
    \item House 2 already serves tea by Clue 3, so the remaining drink, juice, must belong to Carol in House 3.
\end{enumerate}

\medskip
\noindent{\bf Solution.}  
This example also illustrates different types of reasoning steps. The deduction ``Alice cannot live in House 3'' is a \emph{valid} intermediate step. The statement ``Bob drinks coffee'' is \emph{contradictory}, since it violates Clues 2, 3, and 6. A \emph{novel} reasoning step is a valid deduction that is not already implied by previously accepted steps, such as the deduction that Bob must own the cat after combining Clue 1 with the placement of Alice in House 1.

\medskip
\noindent{\bf Complete Solution.}

\begin{table}[h]
\centering
\small
\begin{tabular}{c|c|c|c}
\toprule
\textbf{House} & \textbf{Person} & \textbf{Pet} & \textbf{Drink} \\
\midrule
1 & Alice & Dog  & Coffee \\
2 & Bob   & Cat  & Tea \\
3 & Carol & Fish & Juice \\
\bottomrule
\end{tabular}
\end{table}
\subsection{AR-LSAT}
\label{app:ar-lsat}

AR-LSAT is a benchmark derived from the analytical reasoning section of the Law School Admission Test. Each instance consists of a passage describing a structured scenario, a question, and multiple answer options. Solving the task requires identifying the entities involved, interpreting the governing rules, and reasoning over their consequences to determine the correct answer. The benchmark was introduced to study analytical reasoning over text using questions collected from LSAT exams from 1991 to 2016, and prior work shows that these problems are challenging for neural models because they require explicit reasoning over constraints rather than shallow lexical matching~\citep{zhong2022analytical}. 

A key feature of AR-LSAT is that its instances cover multiple reasoning types. In this work, we focus on the three major categories commonly used in prior studies: \emph{ordering}, \emph{grouping}, and \emph{assignment}. Ordering problems require arranging entities subject to precedence or positional rules. Grouping problems require partitioning entities into valid subsets under compatibility constraints. Assignment problems require mapping entities to roles, slots, or resources while satisfying exclusivity and dependency constraints. This diversity makes AR-LSAT a useful complement to ZebraPuzzles: unlike ZebraPuzzles, which emphasize fully specified grid-style assignments, AR-LSAT is more text-centric and question-oriented, requiring the model to reason over partial consequences of a rule system~\citep{zhong2022analytical}. 

AR-LSAT is also well matched to solver-guided reasoning. Although the input is fully textual, the scenario can often be converted into a structured representation involving participants, positions, groups, or assignments together with logical constraints. Prior neurosymbolic work has shown that combining LLMs with symbolic solvers can substantially improve performance on AR-LSAT, highlighting the usefulness of explicit formalization for solving such problems~\citep{pan2023logiclm}.

\noindent{\bf Example.}
Consider a simple AR-LSAT-style ordering problem involving three presentations:
\[
\{A, B, C\}.
\]
Each presentation must be scheduled in exactly one of the three positions: first, second, and third.

\medskip
\noindent{\bf Rules.}
\begin{enumerate}
    \item $A$ must occur before $B$.
    \item $C$ cannot be first.
    \item $B$ cannot be third.
\end{enumerate}

\medskip
\noindent{\bf Question.}
Which of the following must be true?
\begin{enumerate}
    \item $A$ is first.
    \item $B$ is third.
    \item $C$ is second.
    \item $A$ is third.
\end{enumerate}

\medskip
\noindent{\bf Reasoning Steps.}  
To solve this problem, the following intermediate reasoning steps can be derived.
\begin{enumerate}
    \item Since $C$ cannot be first, either $A$ or $B$ must be first.
    \item Because $A$ must occur before $B$, $B$ cannot be first.
    \item Therefore, $A$ must be first.
    \item Since $B$ cannot be third, $B$ must be second.
    \item Consequently, $C$ must be third.
\end{enumerate}

\medskip
\noindent{\bf Final Answer.}  
The statements that must be true are:
\[
A \text{ is first}, \quad B \text{ is second}, \quad C \text{ is third}.
\]
Therefore, among the answer choices above, the correct option is:
\[
\boxed{A \text{ is first}}
\]

\medskip
This example illustrates step-level reasoning in AR-LSAT. The deduction ``$A$ must be first'' is a \emph{valid} intermediate step. The statement ``$B$ is first'' is \emph{contradictory}, since it violates Rule 1. A \emph{novel} reasoning step is a valid deduction that is not already implied by previously accepted steps, such as the inference that ``$B$ must be second'' after combining Rules 1 and 3.

\subsection{Knights and Knaves}
\label{app:knights-knaves}

Knights and Knaves (KnK) consists of logical reasoning puzzles in which
each character is either a knight, who always tells the truth, or a knave,
who always lies~\cite{xie2024memorization}. Each character makes a statement
about one or more characters, and the task is to determine the identity of
every character while ensuring that all statements are logically consistent.
The benchmark contains puzzles with 2--8 characters, requiring increasingly
complex reasoning as the number of characters and dependencies among their
statements grows.

Knights and Knaves is well suited to solver-guided reasoning because each
character's identity can be represented as a Boolean variable and each
statement can be expressed as a logical constraint. A knight's statement
must evaluate to true, whereas a knave's statement must evaluate to false.
The resulting constraints can therefore be checked directly by a symbolic
solver.

\noindent{\bf Example.}
Consider a simple Knights and Knaves puzzle involving three characters:
$\{A,B,C\}$.

\noindent{\bf Statements.}
\begin{enumerate}
    \item A says: ``B is a knave.''
    \item B says: ``A and C are of the same type.''
    \item C says: ``A is a knight.''
\end{enumerate}

\noindent{\bf Reasoning Steps.}
To solve this puzzle, the following intermediate reasoning steps can be derived.
\begin{enumerate}
    \item Suppose A is a knight. Then A's statement is true, so B is a knave.
    \item Since B is a knave, B's statement is false, so A and C are of
    different types.
    \item Since A is a knight, C must therefore be a knave.
    \item However, C's statement that A is a knight is true, contradicting
    the fact that C is a knave.
    \item Therefore, A must be a knave. A's statement is false, so B must
    be a knight.
    \item Since B is a knight, A and C are of the same type. Hence C is
    also a knave.
\end{enumerate}

\noindent{\bf Final Answer.}
A is a knave, B is a knight, and C is a knave.

This example illustrates step-level reasoning in Knights and Knaves.
For example, under the assumption that A is a knight, the deduction that
B is a knave is a valid intermediate step. However, this assumption
eventually leads to a contradiction because C, inferred to be a knave,
makes a true statement. The solver can detect such inconsistencies and
verify the deductions leading to the final assignment.

\section{Additional Experimental Settings}
\label{app:exp}

\subsection{Evaluation Benchmarks}
\label{app:evaluation-datasets}

Table~\ref{tab:dataset_stats} summarizes the data splits used for ZebraLogic,
AR-LSAT, and Knights and Knaves. ZebraLogic contains 1000 puzzles in total,
divided by difficulty into Small, Medium, Large, and Extra-Large categories,
with 250/50/700 instances for training, validation, and testing, respectively.
For AR-LSAT, we construct balanced training and validation sets with 300 and
50 instances, respectively, for each reasoning type: Ordering, Grouping, and
Assignment. The test set follows the official split, containing 112 Ordering,
49 Grouping, and 69 Assignment questions, for a total of 230 evaluation
instances. Knights and Knaves contains 6200 training and 700 test puzzles
spanning 2--8 characters. We sample 300 instances from the training set for RL model training, 50 for validation and evaluate on the complete 700-instance test set.
Overall, these benchmarks cover complementary logical reasoning settings,
including difficulty-controlled constraint puzzles, distinct reasoning types,
and truth-conditional reasoning over multiple characters.
\begin{table*}[t]
\centering
\caption{Data statistics for ZebraLogic, AR-LSAT, and Knights and Knaves.}
\label{tab:dataset_stats}
\small
\setlength{\tabcolsep}{6pt}
\begin{tabular}{l l c c c c}
\toprule
\textbf{Dataset} & \textbf{Category / Split} & \textbf{Train} & \textbf{Validation} & \textbf{Test} & \textbf{Notes} \\
\midrule

\multirow{5}{*}{ZebraLogic}
& Small         & 80  & 12 & 224 & \multirow{5}{*}{1000 total puzzles} \\
& Medium        & 67  & 12 & 196 & \\
& Large         & 52  & 13 & 138 & \\
& Extra-Large   & 51  & 13 & 142 & \\
\cmidrule(lr){2-6}
& Total         & 250 & 50 & 700 & \\

\midrule

\multirow{4}{*}{AR-LSAT}
& Ordering      & 300 & 50 & 112 & \multirow{4}{*}{Official test split} \\
& Grouping      & 300 & 50 & 49  & \\
& Assignment    & 300 & 50 & 69  & \\
\cmidrule(lr){2-6}
& Total         & 900 & 150 & 230 & \\

\midrule

Knights \& Knaves
& Total         & 300 & 50 & 700 & Official test split \\

\bottomrule
\end{tabular}
\end{table*}

\subsection{Evaluation Metrics}
\label{app:evaluation-metrics}

\noindent{\bf (a) Puzzle Acc. (ZebraLogic; Knights and Knaves).}
Puzzle accuracy measures the fraction of puzzles whose complete
predicted assignments match the ground truth. Let $\hat{Y}_i$
and $Y_i$ denote the predicted and ground-truth solutions for
puzzle $i$, respectively. For $N$ test puzzles,
\[
    \mathrm{Puzzle\;Acc}
    = \frac{1}{N}\sum_{i=1}^{N}
      \mathbf{1}\!\left[\hat{Y}_i = Y_i\right].
\]
For ZebraLogic, an exact match requires all cell assignments to be correct; for Knights and Knaves, it requires all character identities to be correct. Neither benchmark awards partial credit under this metric.

\eat{
\noindent{\bf (a) Puzzle Acc.}
Puzzle Acc. measures whether the model solves the entire Zebra Puzzle correctly. Let $\hat{Y}_i$ denote the predicted complete solution for puzzle $i$, and let $Y_i$ denote the corresponding ground-truth solution. For a test set of $N$ puzzles, Puzzle Accuracy is defined as:
\[
\mathrm{Puzzle\;Acc}
=
\frac{1}{N}
\sum_{i=1}^{N}
\mathbf{1}\!\left[\hat{Y}_i = Y_i\right].
\]
This is a strict metric: a puzzle is counted as correct only if all predicted cells match the ground-truth solution.
}

\noindent{\bf (b) Cell Acc. (ZebraLogic) and Person Acc. (Knights \& Knaves).}
These metrics measure the fraction of individual assignments predicted
correctly: cell values in ZebraLogic and character identities in
Knights and Knaves. Let $M_i$ denote the number of assignments in
puzzle $i$, and let $Y_{ij}$ and $\hat{Y}_{ij}$ denote the ground-truth
and predicted values of assignment $j$, respectively. Both metrics
are computed as
\[
\mathrm{Cell/Person\;Acc}
=
\frac{
\sum_{i=1}^{N}\sum_{j=1}^{M_i}
\mathbf{1}\!\left[\hat{Y}_{ij}=Y_{ij}\right]
}{
\sum_{i=1}^{N} M_i
}.
\]
We report this metric as Cell Acc.\ for ZebraLogic and Person Acc.\
for Knights and Knaves. Unlike Puzzle Acc., it gives partial credit
for correctly predicted assignments within a puzzle.

\eat{
\noindent{\bf (b) Cell Acc (ZebraLogic).}
Cell Acc. measures the fraction of individual assignments that are predicted correctly. Let $Y_i$ contain $M_i$ cells, and let $Y_{ij}$ and $\hat{Y}_{ij}$ denote the ground-truth and predicted value of cell $j$ in puzzle $i$, respectively. Cell Accuracy is defined as
\[
\mathrm{CellAcc}
=
\frac{\sum_{i=1}^{N}\sum_{j=1}^{M_i}
\mathbf{1}\!\left[\hat{Y}_{ij} = Y_{ij}\right]}
{\sum_{i=1}^{N} M_i}.
\]
Unlike Puzzle Accuracy, this metric gives partial credit when the model correctly predicts some but not all cells of a puzzle.

\noindent{\bf (e) Person Acc. (Knights \& Knaves).}
Person accuracy measures correctness at the individual-character level.
Using the same notation, it is defined as
\[
\mathrm{Person\ Acc}
=
\frac{
\sum_{i=1}^{N}
\sum_{j=1}^{M_i}
\mathbf{1}\!\left[\hat{y}_{ij}=y_{ij}\right]
}{
\sum_{i=1}^{N} M_i
}.
\]
Unlike Puzzle Acc., which requires the complete assignment for a puzzle to
be correct, Person Acc. gives credit for each correctly predicted character
identity across all test puzzles.}

\noindent{\bf (c) Acc (AR-LSAT).}
For AR-LSAT, each instance is a multiple-choice logical reasoning problem with a single correct answer. Let $\hat{a}_i$ denote the model's predicted option for instance $i$, and let $a_i$ denote the ground-truth option. Accuracy is defined as
\[
\mathrm{Acc}
=
\frac{1}{N}
\sum_{i=1}^{N}
\mathbf{1}\!\left[\hat{a}_i = a_i\right].
\]
We report accuracy separately for Ordering, Grouping, and Assignment problems, as well as overall accuracy across all AR-LSAT test instances.
Overall accuracy is the weighted mean of the accuracies for Ordering, Grouping, and Assignment.

\begin{table}[t!]
\centering
\caption{\OurMODEL{} performance aggregated by puzzle category using the predefined size groups. These results are reported using Qwen3-4B-Thinking.}
\label{tab:categorywise_metrics}
\small
\setlength{\tabcolsep}{6pt}
\resizebox{0.75\linewidth}{!}{%
\begin{tabular}{l c c c c c}
\toprule
\textbf{Category} & \textbf{\#Samples} & \textbf{Cell-Acc.} & \textbf{SAT-rate} & \textbf{\#Novel-steps} & \textbf{Puzzle Acc.} \\
\midrule
Small        & 224 & 0.941 & 0.924 & 2.81 & 0.915 \\
Medium       & 196 & 0.922 & 0.908 & 4.66 & 0.888 \\
Large        & 138 & 0.882 & 0.884 & 4.92 & 0.783 \\
Extra-Large  & 142 & 0.405 & 0.641 & 2.08 & 0.204 \\
\bottomrule
\end{tabular}%
}
\end{table}

\section{Additional Experimental Results}
\label{app:exp_addl}

To better understand the gains achieved by~\OurMODEL{}, we conduct several additional analyses. These include: (i) Scaling behavior of~\OurMODEL{} across different puzzle sizes; (ii) Z3-solver SAT-rate across training epochs; (iii) Error Analysis, and (iv) Qualitative Case studies.

\begin{table*}[hb!]
\centering
\caption{Test metrics by inferred puzzle size. The puzzle size is extracted from \texttt{pid} as A×B, and each size is mapped to a predefined difficulty category. All results are obtained using Qwen3-4B-Thinking.}
\label{tab:sizewise_metrics}
\small
\setlength{\tabcolsep}{6pt}
\resizebox{0.60\textwidth}{!}{
\begin{tabular}{l l c c c c c}
\toprule
\textbf{Category} & \textbf{Size} & \textbf{\#Samples} & \textbf{Cell Acc.} & \textbf{SAT-rate} & \textbf{\#Novel-steps} & \textbf{Puzzle Acc.} \\
\midrule

\multirow{8}{*}{Small}
& 2x2   & 23 & 1.000 & 0.957 & 0.96 & 1.000 \\
& 2x3   & 27 & 1.000 & 1.000 & 2.15 & 1.000 \\
& 2x4   & 31 & 0.935 & 0.871 & 2.84 & 0.935 \\
& 2x5   & 26 & 0.904 & 0.923 & 2.96 & 0.846 \\
& 2x6   & 29 & 0.877 & 0.828 & 2.83 & 0.793 \\
& 3x2   & 33 & 0.970 & 0.939 & 3.24 & 0.970 \\
& 3x3   & 29 & 0.980 & 1.000 & 3.97 & 0.931 \\
& 4x2   & 26 & 0.859 & 0.885 & 3.12 & 0.846 \\
\cmidrule(lr){2-7}
& Total & 224 & 0.941 & 0.924 & 2.81 & 0.915 \\

\midrule

\multirow{8}{*}{Medium}
& 3x4   & 26 & 0.964 & 1.000 & 5.69 & 0.923 \\
& 3x5   & 29 & 0.931 & 0.931 & 6.62 & 0.931 \\
& 3x6   & 28 & 0.905 & 0.929 & 3.43 & 0.857 \\
& 4x3   & 30 & 0.883 & 0.900 & 4.13 & 0.867 \\
& 4x4   & 27 & 0.896 & 0.815 & 4.85 & 0.852 \\
& 5x2   & 29 & 0.931 & 0.931 & 3.28 & 0.931 \\
& 6x2   & 27 & 0.947 & 0.852 & 4.70 & 0.852 \\
\cmidrule(lr){2-7}
& Total & 196 & 0.922 & 0.908 & 4.66 & 0.888 \\

\midrule

\multirow{6}{*}{Large}
& 4x5   & 29 & 0.864 & 0.828 & 3.86 & 0.759 \\
& 4x6   & 26 & 0.801 & 0.923 & 5.08 & 0.692 \\
& 5x3   & 29 & 0.984 & 0.931 & 4.14 & 0.966 \\
& 5x4   & 31 & 0.836 & 0.871 & 5.94 & 0.677 \\
& 6x3   & 23 & 0.931 & 0.870 & 5.70 & 0.826 \\
\cmidrule(lr){2-7}
& Total & 138 & 0.882 & 0.884 & 4.92 & 0.783 \\

\midrule

\multirow{6}{*}{X-Large}
& 5x5   & 28 & 0.540 & 0.714 & 2.54 & 0.393 \\
& 5x6   & 30 & 0.421 & 0.667 & 2.27 & 0.200 \\
& 6x4   & 24 & 0.669 & 0.708 & 3.12 & 0.458 \\
& 6x5   & 27 & 0.201 & 0.704 & 1.11 & 0.000 \\
& 6x6   & 33 & 0.253 & 0.455 & 1.58 & 0.030 \\
\cmidrule(lr){2-7}
& Total & 142 & 0.405 & 0.641 & 2.08 & 0.204 \\

\bottomrule
\end{tabular}
}
\end{table*}

\subsection{Scaling Behavior of~\OurMODEL{} Across Puzzle Sizes}
\label{app:scaling_puzzle_size}
Table~\ref{tab:categorywise_metrics} and
Table~\ref{tab:sizewise_metrics} report the performance of~\OurMODEL{} using Qwen3-4B-Thinking as LLM, across puzzle sizes and their corresponding difficulty categories. The results show a clear size-dependent degradation trend. On Small puzzles, the model performs strongly, achieving \(0.941\) Cell Acc., \(0.924\) SAT-rate, and \(0.915\) Puzzle Acc., which indicates that it can usually recover both locally correct cell assignments and fully correct global solutions. Performance remains relatively high on Medium puzzles, although the average number of novel reasoning steps increases from \(2.81\) to \(4.66\), suggesting that these instances require more active multi-step inference. On Large puzzles, Cell Acc. and SAT-rate remain reasonably stable, but Puzzle Acc. drops more noticeably to \(0.783\), indicating that exact end-to-end solving becomes harder even when many local assignments are still correct. The most severe degradation appears on X-Large puzzles, where Cell Acc. falls to \(0.405\) and Puzzle Acc. to \(0.204\). Notably, the SAT-rate on X-Large puzzles remains substantially higher than Puzzle Acc. (\(0.641\) vs.\ \(0.204\)), suggesting that the model often produces partially solver-consistent structures without fully matching the ground-truth solution. Moreover, the decline in \#Novel-steps on X-Large puzzles implies that reasoning productivity itself begins to break down under the most complex settings.

\subsection{Z3-Solve SAT rate across epochs}
\label{sec:sat_rate}

\begin{figure*}[h!]
    \centering
    \vspace{-5pt}
    \includegraphics[width=0.65\linewidth]{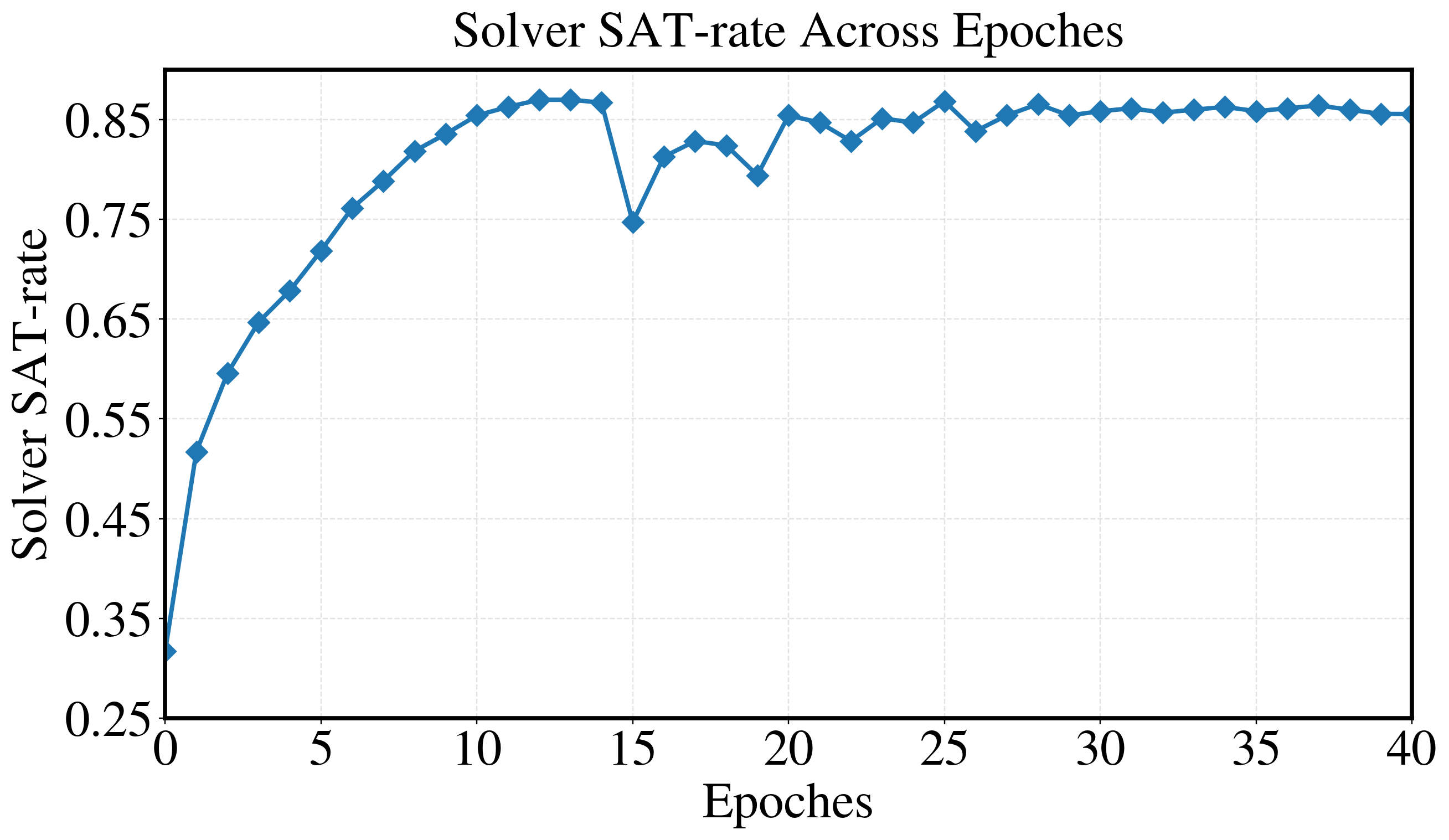}
    \vspace{-5pt}
    \caption{Z3 Solver SAT-rate across training epochs.}
    \vspace{-2pt}
    \label{fig:solver_sat_rate}
\end{figure*}

Figure~\ref{fig:solver_sat_rate} shows the evolution of the solver SAT-rate for test instances of the ZebraLogic benchmark across different epochs using Qwen3-4B-Thinking LLM. This metric measures the proportion of generated formal reasoning trajectories that can be translated into satisfiable Z3 constraints under the underlying puzzle formulation. The SAT-rate increases rapidly during the early stages of training, indicating that the model quickly learns to produce solver-compatible reasoning steps. It then stabilizes around the mid-\(0.8\) range with only minor fluctuations, suggesting convergence toward a relatively stable regime of logically consistent reasoning. Since satisfiable Z3 formulations reflect whether the generated intermediate deductions preserve consistency at the constraint level, a higher SAT-rate provides evidence that the model is producing more reliable and solver-grounded reasoning trajectories over time.

\begin{table}[t!]
\centering
\caption{Error analysis with respect to puzzle size for ZebraLogic benchmark. Error count refers to instances with \(\texttt{Puzzle Acc.}=0\).}
\label{tab:error_wrt_size}
\small
\setlength{\tabcolsep}{6pt}
\resizebox{0.55\linewidth}{!}{%
\begin{tabular}{l c c c c}
\toprule
\textbf{Category} & \textbf{\#Total} & \textbf{\#Correct} & \textbf{\#Incorrect} & \textbf{Error Rate} \\
\midrule
Small       & 224 & 205 & 19  & 0.085 \\
Medium      & 196 & 174 & 22  & 0.112 \\
Large       & 138 & 108 & 30  & 0.217 \\
X-Large     & 142 & 29  & 113 & 0.796 \\
\bottomrule
\end{tabular}%
}
\end{table}

\begin{table*}[h!]
\centering
\caption{Statistics of incorrect cases for ZebraLogic benchmark. Counts are computed over instances with \(\texttt{Puzzle Acc.}=0\).}
\label{tab:error_reasons}
\small
\setlength{\tabcolsep}{6pt}
\resizebox{0.60\textwidth}{!}{%
\begin{tabular}{l c c c c c}
\toprule
\textbf{Reason} & \textbf{\#Total} & \textbf{Small} & \textbf{Medium} & \textbf{Large} & \textbf{X-Large} \\
\midrule
\texttt{Solver-Consistent}       & 84 & 6  & 6  & 16 & 56 \\
\texttt{Missing output}          & 40 & 11 & 11 & 1  & 17 \\
\texttt{Header Mismatch}         & 35 & 1  & 4  & 8  & 22 \\
\texttt{Solver-Inconsistent}     & 25 & 1  & 1  & 5  & 18 \\
\texttt{Format Failure}          & 9  & 0  & 0  & 0  & 9  \\
\bottomrule
\end{tabular}%
}
\end{table*}

\subsection{Error Analyses}
\label{app:error_analyses}

Table~\ref{tab:error_wrt_size} reports the distribution of correct and incorrect cases across puzzle-size categories, where an error is defined as an instance with \(\texttt{Puzzle Acc.}=0\). The results show a clear and strong size-dependent degradation pattern. On Small and Medium puzzles, the model performs reliably, with low error rates of \(0.085\) and \(0.112\), respectively, indicating that most instances in these categories are solved correctly. The error rate nearly doubles on Large puzzles to \(0.217\), suggesting that exact solution recovery becomes noticeably harder as the combinatorial structure grows. The sharpest breakdown occurs on X-Large puzzles, where the error rate rises dramatically to \(0.796\): only \(29\) out of \(142\) instances are solved correctly, while \(113\) are incorrect. This trend indicates that puzzle size is a major driver of failure, and that the model’s reasoning and decoding pipeline remain robust on smaller instances but degrade substantially under the complexity of the hardest settings.

\begin{table*}[h!]
\centering
\caption{Different types of Error Cases for ZebraLogic Benchmark.}
\label{tab:qualitative_error_examples}
\scriptsize
\setlength{\tabcolsep}{3pt}
\renewcommand{\arraystretch}{1.12}

\resizebox{\textwidth}{!}{%
\begin{tabular}{>{\raggedright\arraybackslash}p{2.5cm}
                >{\raggedright\arraybackslash}p{3.6cm}
                >{\raggedright\arraybackslash}p{4.8cm}
                >{\raggedright\arraybackslash}p{7.2cm}}
\toprule
\textbf{Reason} & \textbf{Key signals} & \textbf{1-line prediction snippet} & \textbf{Observation} \\
\midrule

\multirow{2}{*}{\texttt{Missing output}}
& Cell Acc.\(=0\), SAT\(=0\), Puzzle Acc.\(=0\)
& WRONG OUTPUT FORMAT
& The model returns no usable prediction at all, so the failure is caused by blank or unextractable output rather than reasoning. \\

\midrule

\multirow{2}{*}{\texttt{Header Mismatch}}
& \multirow{2}{*}{\(\texttt{Format\_Check}=\texttt{False}\)}
& header: [House, Name, FavoriteSport] \(\rightarrow\) [House, Name, Sport]
& Mis-match in header schema output fields. \\

\midrule

\multirow{2}{*}{\texttt{Solver-Consistent}}
& \multirow{2}{*}{SAT\(=1\), Puzzle Acc.\(=0\)}
& rows: [1, Eric, colonial, lilies], [2, Arnold, colonial, lilies], \dots
& The output remains solver-consistent, but duplicated attribute assignments across houses make it non-exact.\\

\midrule

\multirow{2}{*}{\texttt{Solver-Inconsistent}}
& \multirow{2}{*}{SAT\(=0\), Puzzle Acc.\(=0\)}
& houses 1--2 both use Mother=Holly, Child=Alice, Animal=cat
& \texttt{SAT=0}. Many local assignments are plausible, but repeated values violate the global one-to-one puzzle constraints.\\

\midrule

\multirow{2}{*}{\texttt{Format Failure}}

& \multirow{2}{*}{\(\texttt{Format\_Check}=\texttt{False}\)}
& Header has 7 fields, but each predicted row contains only 6 values
& The output has the shape of a table, but every row is missing one attribute column, so evaluation collapses to zero matched cells.\\

\bottomrule
\end{tabular}%
}
\end{table*}

Tables~\ref{tab:error_reasons} provide a complementary quantitative analysis of the incorrect cases to different error types (types are illustrated in Table~\ref{tab:qualitative_error_examples}). It shows that the dominant failure mode is \texttt{Solver-Consistent} prediction, where the model produces outputs that remain globally satisfiable but do not exactly match the gold solution; notably, this category becomes especially prominent on X-Large puzzles. The next major error sources are \texttt{Missing output} and \texttt{Header Mismatch}, indicating that a substantial fraction of failures arise not only from reasoning errors, but also from generation or formatting issues. In contrast, \texttt{Solver-Inconsistent} and \texttt{Format Failure} cases are less frequent overall, but they become much more visible on harder instances, suggesting that increasing puzzle complexity amplifies both global inconsistency and structural output breakdown.

\begin{table*}[h!]
\centering
\caption{Case study on shared puzzles with~\OurMODEL{} compared against (\texttt{--N}) and (\texttt{--NC}) variants. We used Qwen3-4B-Thinking for this analysis.}
\label{tab:case_study_same_puzzle}
\small
\setlength{\tabcolsep}{4pt}
\resizebox{0.65\textwidth}{!}{%
\begin{tabular}{l l c c c c c c}
\toprule
\textbf{Puzzle ID} & \textbf{System} & \textbf{Puzzle Acc.} & \textbf{SAT} & \textbf{\#Steps} & \textbf{\#Parsed} & \textbf{\#Valid} & \textbf{\#Contradictions} \\
\midrule

\multirow{3}{*}{\texttt{4x5-16}}
& \OurMODEL{} & 1.0 & 1.0 & 22  & 11 & 11 & 0 \\
& \texttt{--N}  & 0.0 & 1.0 & 85  & 40 & 22 & 18 \\
& \texttt{--NC} & 0.0 & 1.0 & 144 & 34 & 31 & 3 \\
\midrule

\multirow{3}{*}{\texttt{4x6-0}}
& \OurMODEL{} & 1.0 & 1.0 & 20 & 10 & 10 & 0 \\
& \texttt{--N}  & 0.0 & 1.0 & 88 & 42 & 26 & 16 \\
& \texttt{--NC} & 0.0 & 0.0 & 0  & 0  & 0  & 0 \\
\midrule

\multirow{3}{*}{\texttt{4x4-36}}
& \OurMODEL{} & 1.0 & 1.0 & 38 & 19 & 19 & 0 \\
& \texttt{--N}  & 0.0 & 1.0 & 54 & 24 & 18 & 6 \\
& \texttt{--NC} & 0.0 & 1.0 & 84 & 24 & 18 & 6 \\
\midrule

\multirow{3}{*}{\texttt{6x3-7}}
& \OurMODEL{} & 1.0 & 1.0 & 16 & 8  & 8  & 0 \\
& \texttt{--N}  & 0.0 & 1.0 & 60 & 28 & 16 & 12 \\
& \texttt{--NC} & 0.0 & 0.0 & 0  & 0  & 0  & 0 \\
\bottomrule
\end{tabular}%
}
\end{table*}

\subsection{Qualitative Case study}
\label{app:case_study}
\begin{table*}[h!]
\centering
\caption{Case study on a matched puzzle. We compare the parsed-only reasoning traces produced by \OurMODEL{}, \texttt{--N}, and \texttt{--NC} on the same puzzle. We used Qwen3-4B-Thinking for this analysis.}
\label{tab:case_study_example}
\scriptsize
\setlength{\tabcolsep}{4pt}
\renewcommand{\arraystretch}{1.15}

\begin{tabularx}{0.70\textwidth}{
    >{\centering\arraybackslash}p{1.8cm}
    >{\raggedright\arraybackslash}X
    >{\centering\arraybackslash}p{4.2cm}
}
\toprule
\textbf{System} & \textbf{Parsed-only reasoning trace} & \textbf{Outcome} \\
\midrule

\multirow{6}{*}{\OurMODEL{}}
&
\texttt{(1) House 2 = Eric;} \newline
\texttt{(2) Eric \(\rightarrow\) colonial;} \newline
\texttt{(3) House 3 \(\neq\) lilies;} \newline
\texttt{(4) House 1 = Arnold;} \newline
\texttt{(5) Arnold \(\rightarrow\) roses;} \newline
\texttt{(6) Therefore House 4 = lilies;} \newline
\texttt{(7) \dots}
&
\multirow{6}{*}{Exact solution; short, contradiction-free trace.} \\

\midrule

\multirow{6}{*}{\texttt{--N}}
&
\texttt{(1) House 2 = Eric;} \newline
\texttt{(2) Eric \(\rightarrow\) colonial;} \newline
\texttt{(3) House 2 = Arnold;} \newline
\texttt{(4) Arnold \(\rightarrow\) lilies;} \newline
\texttt{(5) House 4 = lilies;} \newline
\texttt{(6) House 1 = Eric;} \newline
\texttt{(7) \dots}
&

\multirow{6}{*}{Solver-consistent but duplicated assignments.} \\

\midrule

\multirow{6}{*}{\texttt{--NC}}
&
\texttt{(1) House 2 = Eric;} \newline
\texttt{(2) Eric \(\rightarrow\) colonial;} \newline
\texttt{(3) House 1 = Arnold;} \newline
\texttt{(4) Arnold \(\rightarrow\) colonial;} \newline
\texttt{(5) House 3 = lilies;} \newline
\texttt{(6) House 4 = lilies;} \newline
\texttt{(7) \dots}
&
\multirow{6}{*}{Long, noisy trace with repeated attributes.} \\

\bottomrule
\end{tabularx}
\end{table*}
Here we provide a more detailed matched-puzzle analysis that combines quantitative trace statistics with qualitative inspection of the reasoning trajectories. From the quantitative perspective, Table~\ref{tab:case_study_same_puzzle} shows that, on shared validation puzzles, \OurMODEL{} consistently solves the instance with fewer steps and zero contradictions, while both (\texttt{--N}) and (\texttt{--NC}) fail on the same puzzles despite often generating substantially longer traces. This confirms that the benefit of \OurMODEL{} is not simply higher final accuracy, but also more efficient reasoning on identical instances. From the qualitative perspective, Table~\ref{tab:case_study_example} illustrates parsed-only reasoning steps for (\texttt{--N}) and (\texttt{--NC}) model variants. \OurMODEL{} tends to maintain a compact, monotonic deduction chain, while the (\texttt{--N}) and (\texttt{--NC}) repeatedly violate global consistency. A recurring error is contradictory reassignment, where a trace first states ``House 2 = Eric'' and later also states ``House 2 = Arnold.'' Another common pattern is duplicated attribute placement, where the same value, such as a flower or role, is attached to multiple houses even though the puzzle enforces uniqueness. 

Beyond the illustrative example in the table, we also observe cases of unstable clue propagation, where a baseline first makes a plausible local deduction but later reuses the same person or attribute incompatibly. For example, a trace may infer ``House 2 = Eric'' and ``Eric \(\rightarrow\) colonial,'' but later also assign ``Arnold \(\rightarrow\) colonial,'' or place ``lilies'' in multiple houses despite the uniqueness constraint.
Overall, this case study strengthens the main results by showing that solver-guided process supervision with \emph{novel reasoning steps} improves both the efficiency of reasoning and the internal consistency of the reasoning trajectory itself.

\section{Limitations}
\label{app:limitations}

Our work poses the following limitations. First,~\OurMODEL{} relies on the availability of a solver-compatible symbolic representation. Although this is natural for structured logical reasoning tasks, such as Zebra-style puzzles and AR-LSAT, it limits direct applicability to problems whose premises and intermediate deductions can be reliably translated into formal constraints. For more open-ended reasoning domains, the required symbolic interface may be difficult to define or may introduce additional parsing errors.

Second, the framework depends on the quality of the Parsed Syntactic Premises generated by the policy LLM. If the parsed premises are incomplete, incorrect, or solver-incompatible, the resulting solver state may become unsatisfiable or fail to preserve the intended problem semantics. In such cases, the downstream process rewards become less informative, since the symbolic verifier can only assess reasoning relative to the constructed solver state rather than the original natural-language problem.

Third, our notion of process quality is intentionally centered on solver-verifiable properties, such as validity, novelty, contradiction, and consistency with the final solution. While these signals are effective for structured logic problems, they do not capture all desirable aspects of reasoning, such as abstraction, explanation quality, or human interpretability beyond the constrained symbolic setting. As a result, the framework may favor reasoning that is solver-compatible without necessarily being the most concise or natural from a human perspective.

\clearpage

\section{Prompts}
\label{app:prompts}

\subsection{ZebraLogic Prompt}
\label{app:zebra_prompt}

\begin{promptbox}
\begin{Verbatim}[breaklines=true,breakanywhere=true,fontsize=\footnotesize]
"""
You are an expert logic puzzle solver.

You are given:
(i) one logic puzzle_text written in plain English,
(ii) solution_header that lists the attribute names used in the puzzle, and
(iii) a dictionary of attribute_values specifying the complete and exclusive set of allowed values for each attribute.

All values appearing in syntactic_clues, reasoning, and the final solution MUST be drawn from attribute_values and interpreted as entity tokens representing unknown house positions.

Your task is to construct a fully consistent, solver-verifiable solution by generating the following FIVE fields:
1) n_houses — the total number of houses in the puzzle.
2) attribute_values — returned exactly as given, without modification.
3) syntactic_clues — a normalized, Z3-style textual encoding of each clue.
4) reasoning — interleaved reasoning consisting of natural-language explanations and syntactic (solver-checkable) deduction steps.
5) solution — the final house-by-house assignment derived exclusively from syntactic_clues, and syntactic reasoning steps (S1..Sk).


You MUST return the result STRICTLY as a single valid JSON object wrapped inside:
<answer>...</answer>

No additional text, commentary, or formatting outside the <answer> block is permitted.


================================================================================
CRITICAL FORMAT REQUIREMENTS
================================================================================
- Output MUST contain ONLY ONE <answer>...</answer> block and NOTHING ELSE.
- Do NOT include extra text, markdown, explanations, or code fences.
- Inside <answer>...</answer>, the content MUST be a single valid JSON object.
- The JSON object MUST have exactly FIVE top-level keys, spelled EXACTLY:
    "n_houses",
    "attribute_values",
    "syntactic_clues",
    "reasoning",
    "solution"
- Do NOT add any other keys.

================================================================================
NORMALIZATION RULES
================================================================================
- Use underscores instead of spaces in VALUES (e.g., grilled_cheese, very_short).
- Attribute names MUST match the solution_header exactly (case-sensitive), e.g., Name, Animal, Occupation, Sport, Height, etc.
- House numbers are integers 1..N.
- Convert ordinals to integers: first=1, second=2, third=3, fourth=4, fifth=5, sixth=6, etc.
- Do not invent values. Every value must be mapped to its canonical token (in attribute_values) and selected from the list of allowed attribute_values (after normalization).
 - Example: If puzzle says “september” and attribute_values contains "sept", output "sept" (not september).
 - Example: If puzzle says “sept” and attribute_values contains "september", output "september"
- If the clue mentions a bare person name (e.g., "Arnold"), treat it as Name=Arnold.
- If the clue uses a descriptor like "cat lover", "dog owner", "coffee drinker", map it to the matching token in attribute_values.

================================================================================
1) DOMAIN OUTPUT (MANDATORY)
================================================================================
- "n_houses" MUST be an integer N equal to the number of houses in the puzzle.
attribute_values immutability rule:
- The "attribute_values" object MUST be returned exactly as provided in the input.
- It must be identical:
  - Same attribute keys
  - Same ordering of keys
  - Same ordering of values within each list
  - Same casing and spelling
- Do NOT normalize, rename, reorder, add, or remove anything in "attribute_values".
- Normalization rules apply ONLY to syntactic_clues, reasoning, and solution — NOT to attribute_values.

================================================================================
2) syntactic_clues (MANDATORY, TEXTUAL CONSTRAINTS — NOT PREDICATES)
================================================================================
We do NOT use predicate-style DSL for clues.
Instead, each clue MUST be rewritten as a single-line *syntactic constraint statement* in a Z3-like textual form.

Rules:
- "syntactic_clues" MUST be a list of strings.
- For each clue, the selected tokens must be mapped to one of the values defined in attribute_values.
 - Example: If the clue says “sept” and attribute_values contains "september", use "september"; if attribute_values contains "sept", use "sept".
- There MUST be exactly one entry per clue, in the same order as the clues.
- Each entry MUST be exactly 1 line and end with a period.
- Each entry MUST start with the clue id prefix: "C<i>: ".
- Use ONLY these syntactic operators in the clue text:
    ==   (same house / equivalence)
    !=   (not same house)
    <    (somewhere left of)
    >    (somewhere right of)
    + k == (k is a positive integer, e.g., 1 for immediately left, 2 for one house between, 3 for two houses between)
    == H  (fixed house index, where H is an integer)
- Use bare normalized tokens (no quotes) for values (e.g., Arnold, engineer, very_short).
- When a clue states a specific house like "in the fifth house", encode as: <token> == 5
  Example: "The lawyer is in the fifth house." -> "C9: lawyer == 5."
- When a clue states "directly left of", encode as: A + 1 == B
  Example: "baseball is directly left of engineer" -> "C12: baseball + 1 == engineer."
- When a clue states "one house between", encode as: A + 2 == B
  Example: "There is one house between Eric and the bird keeper" -> "C12: Eric + 2 == bird_keeper."
  Example: "There is one house between Arnold and Peter" -> "C12: Arnold + 2 == Peter."
- When a clue states "two houses between", encode as: A + 3 == B
  Example: "There are two houses between Eric and Arnold" -> "C12: Eric + 3 == Arnold."
- When a clue states "person who has", encode as: A == B
  Example: "The person whose mother's name is Holly is the person who has black hair" -> "C12: Holly == black."
- When a clue states "one house between the person who has", encode as: A + 2 == B
  Example: "There is one house between the person who has black hair and Eric" -> "C12: black + 2 == Eric."
- When a clue states "next to each other", encode it as: Or(A == B + 1, A == B - 1)
  Example: "The person who prefers city breaks and Alice are next to each other" -> C12: "Or(city_breaks == Alice + 1, city_breaks == Alice - 1)."
- When a clue states "somewhere to the left of", encode as: A < B
- When a clue states "somewhere to the right of", encode as: A > B
- When a clue states "X is the Y", encode as: X == Y

IMPORTANT:
- The goal is to produce constraints that resemble:
  s.add(<left> <op> <right>)
  but you must NOT write "s.add(...)".
  Only output the inner constraint as text.

================================================================================
3) reasoning (MANDATORY — INTERLEAVED NATURAL + SYNTACTIC)
================================================================================
- "reasoning" MUST be a list of strings.
- Each entry MUST be exactly 1 sentence and end with a period.
- Reasoning MUST be interleaved:
    Odd-numbered entries: Natural-language reasoning.
    Even-numbered entries: Syntactic reasoning step (Z3-like statement).
- Natural-language entries should explain the deduction in plain English.
- Syntactic entries should encode the *newly deduced fact* as a Z3-like statement.
- Tokens in Syntactic entries should encode the *mapped* to values in "attribute_values".

Syntactic entry format:
- Every syntactic entry MUST start with "S<k>: " and MUST end with a period.
- <k> starts at 1 and increments by 1 for each syntactic step only (S1, S2, S3, ...).
- The syntactic constraint MUST be solver-verifiable and may use ONLY:
  ==, !=, <, >, + d ==, Not(...), And(...), Or(...)

- Each syntactic step MUST be written in the exact form: S<k>
    
  Atomic operators:
    ==        (same house / equivalence)
    !=        (not the same house)
    <         (somewhere to the left of)
    >         (somewhere to the right of)
    + d ==    (directed distance; d is a positive integer)
    == H      (fixed house index; H is an integer in 1..n_houses)

  Boolean operators:
    Not(e)    (negation of a single atomic expression)
    And(e1, e2, ..., en)
    Or(e1, e2, ..., en)

- Boolean operators may ONLY be applied to valid atomic expressions.
- Nested Boolean expressions are allowed but MUST remain solver-verifiable.

Examples of valid INTERLEVED reasoning steps:
    The engineer is assigned to house 2.
    S1: engineer == 2.
    
    Since the engineer occupies house 2, the dog cannot also be in house 2.
    S2: dog != 2.
    
    The cat is immediately to the left of the coffee, so the cat’s house index plus one equals the coffee’s house index.
    S3: cat + 1 == coffee.
    
    The green house appears somewhere to the left of the white house.
    S4: green < white.
    
    The dog is not in the first house.
    S5: Not(dog == 1).
    
    The cat cannot be in house 1 or house 3.
    S6: And(cat != 1, cat != 3).
    
    The milk is located either in house 1 or in house 5.
    S7: Or(milk == 1, milk == 5).
  
Logical validity requirement:
- Every syntactic step MUST be logically entailed by the syntactic_clues plus any earlier syntactic steps.
- Do NOT output syntactic steps that merely restate a clue unless they are required as part of the deduction chain.

================================================================================
4) solution (MANDATORY TABLE)
================================================================================
- "solution" MUST be in tabular form with:
  - "header": a list of column names
  - "rows": a list of rows, each row being a list of strings matching the header order
- The header MUST include "House" and then all attribute columns from the puzzle text.
- The rows MUST list houses in increasing order from 1..N.
- All solution values MUST be normalized with underscores.


"""
\end{Verbatim}
\end{promptbox}

\subsection{Prompt Template for AR-LSAT (Grouping Problems)}
\label{app:prompt-arlsat}

\begin{promptbox}
\begin{Verbatim}[breaklines=true,breakanywhere=true,fontsize=\footnotesize]

"""
You are an expert AR-LSAT grouping-game solver.

You are given:
(i) one AR-LSAT grouping passage written in plain English,
(ii) one question about that passage,
(iii) a question_type label,
(iv) a dictionary of answer options,
and optionally
(v) metadata such as tags or entity hints if available.

This prompt is ONLY for GROUPING problems.

Your task is to parse the grouping problem into a solver-oriented logical representation and determine the correct answer by generating the following EIGHT fields:
1) problem_type — must be "grouping".
2) world_model — entities, groups, and structural assumptions.
3) rules — formalized passage rules only.
4) facts — question-specific temporary conditions only.
5) question_semantics — how the options must be evaluated using the provided question_type.
6) options — formalized answer options.
7) reasoning — interleaved natural-language reasoning and formal solver-oriented steps.
8) solution — the final selected answer option.

You MUST return the result STRICTLY as a single valid JSON object wrapped inside:
<answer>...</answer>

================================================================================
CRITICAL FORMAT REQUIREMENTS
================================================================================
- Output MUST contain ONLY ONE <answer>...</answer> block and NOTHING ELSE.
- JSON MUST contain EXACTLY the 8 required keys.
- "problem_type" MUST be exactly "grouping".
- All formal expressions MUST be strings.
- No markdown, no code, no explanation outside <answer>.

================================================================================
NORMALIZATION RULES FOR GROUPING
================================================================================
- Use concise symbolic tokens.
- Preserve entity names exactly (A, B, C, etc.).
- Use group labels exactly as defined (X, Y, Z, Shelf1, Shelf2, etc.).
- Represent assignments using:

    Assign(entity, group)

- Do NOT use numeric positions unless explicitly required.
- Each entity must belong to exactly one group.

================================================================================
PARSING INSTRUCTIONS FOR GROUPING
================================================================================

Construct world_model:
- Extract entities.
- Extract groups.
- Add assumptions:
    each entity belongs to exactly one group,
    groups are mutually exclusive.

Parse rules:
- Use ONLY passage rules.
- Do NOT include question facts here.

Parse facts:
- Add temporary conditions from question.

Parse question semantics:
- Use provided question_type.

Parse options:
- Represent using Assign(...) expressions.

================================================================================
ALLOWED FORMAL OPERATORS FOR GROUPING
================================================================================

Assignment:
    Assign(A, X)

Equality:
    Assign(A, X) == Assign(B, X)
    Assign(A, X) != Assign(B, X)

Boolean:
    And(...)
    Or(...)
    Not(...)
    Implies(...)

Counting:
    AtLeast(k, ...)
    AtMost(k, ...)
    Exactly(k, ...)

Solver:
    Sat(...)
    Unsat(...)

================================================================================
GROUPING EXPRESSION GUIDE
================================================================================

A is in group X:
    Assign(A, X)

A and B are in same group:
    Assign(A, X) == Assign(B, X)

A and B are in different groups:
    Assign(A, X) != Assign(B, X)

If A is in X then B is in Y:
    Implies(Assign(A, X), Assign(B, Y))

Exactly k elements in X:
    Exactly(k, Assign(A, X), Assign(B, X), ...)

================================================================================
REASONING REQUIREMENTS FOR GROUPING PROBLEMS
================================================================================
- "reasoning" MUST be a list of strings.
- Each entry MUST be exactly one sentence and end with a period.
- Reasoning MUST be interleaved:
    Odd-numbered entries: natural-language reasoning.
    Even-numbered entries: formal solver-oriented step.
- Natural-language entries must explain why the next formal step follows from rules, facts, earlier steps, or option testing.
- Formal entries must encode a newly derived grouping fact, membership restriction, counting restriction, option feasibility result, or forced/impossible group assignment.

Formal step format:
- Every formal step MUST start with "S<k>: " and MUST end with a period.
- <k> starts at 1 and increments by 1 for each formal step only.
- Formal steps must be solver-verifiable and may use ONLY:
    Assign(entity, group), ==, !=, Not(...), And(...), Or(...), Xor(...), Implies(...), AtLeast(k, ...), AtMost(k, ...), Exactly(k, ...), Sat(...), Unsat(...)

Atomic grouping expressions:
    Assign(A, X)               A belongs to group X.
    Not(Assign(A, X))          A does not belong to group X.
    Assign(A, X) == Assign(B, X)   A and B have the same X-membership status.
    Assign(A, X) != Assign(B, X)   A and B have different X-membership status.
    Implies(Assign(A, X), Assign(B, Y))   If A is in X, then B is in Y.

Boolean operators:
    Not(e)
    And(e1, e2, ..., en)
    Or(e1, e2, ..., en)
    Xor(e1, e2)
    Implies(e1, e2)

Counting operators:
    AtLeast(k, Assign(A, X), Assign(B, X), ...)
    AtMost(k, Assign(A, X), Assign(B, X), ...)
    Exactly(k, Assign(A, X), Assign(B, X), ...)

Option-testing operators:
    Sat(Option_A)
    Unsat(Option_A)

Allowed reasoning step types:
- Direct question facts:
    If the question says "If D and F are both on X", a valid step is:
    S1: And(Assign(D, X), Assign(F, X)).

- Forced group membership:
    If rules and facts force G to be in Y, a valid step is:
    S2: Assign(G, Y).

- Group exclusion:
    If A cannot be in X, a valid step is:
    S3: Not(Assign(A, X)).

- Same-group deductions:
    If A and B must be in the same group, a valid step is:
    S4: Assign(A, X) == Assign(B, X).

- Different-group deductions:
    If A and B must be in different groups, a valid step is:
    S5: Assign(A, X) != Assign(B, X).

- Conditional deductions:
    If A being in X would force B into Y, a valid step is:
    S6: Implies(Assign(A, X), Assign(B, Y)).

- Exclusive-choice deductions:
    If exactly one of A or B must be in X, a valid step is:
    S7: Xor(Assign(A, X), Assign(B, X)).

- Capacity or counting deductions:
    If exactly two of A, B, and C must be in X, a valid step is:
    S8: Exactly(2, Assign(A, X), Assign(B, X), Assign(C, X)).

- Option feasibility checks:
    If an option can be extended to at least one full valid grouping, use:
    S9: Sat(Option_D).

- Option impossibility checks:
    If an option cannot be extended to any full valid grouping, use:
    S10: Unsat(Option_A).

Logical validity requirements:
- Every formal step MUST be entailed by rules + facts + earlier accepted formal steps, unless it is an option feasibility step.
- For option feasibility steps:
    Sat(Option_X) means rules + facts + Option_X is satisfiable.
    Unsat(Option_X) means rules + facts + Option_X is unsatisfiable.
- Do NOT output unsupported guesses.
- Do NOT output contradictory steps.
- Do NOT output tautologies such as Or(Assign(A, X), Not(Assign(A, X))).
- Do NOT merely restate every passage rule unless the restatement is needed to connect a deduction.
- Do NOT use ordering operators such as <, >, +1 ==, or position numbers unless the grouping problem explicitly includes ordered groups.
- Do NOT use final answer text as a reasoning step; formal steps must remain solver-oriented.

Examples of valid interleaved grouping reasoning:
    The question condition places both D and F in group X.
    S1: And(Assign(D, X), Assign(F, X)).

    Since F and G must be in different groups and F is in X, G must be in Y.
    S2: Assign(G, Y).

    Since C in X would force D to be in Y, C cannot be in X because D is already in X.
    S3: Not(Assign(C, X)).

    Since E and A must be in different groups, they cannot both be in Y.
    S4: Not(And(Assign(E, Y), Assign(A, Y))).

    Option A forces C into X, which contradicts the derived restriction that C cannot be in X.
    S5: Unsat(Option_A).

    Option D can be extended to a complete valid grouping.
    S6: Sat(Option_D).
    

================================================================================
SOLUTION REQUIREMENTS
================================================================================

"solution": {
  "selected_option": "X"
}

================================================================================
OUTPUT SCHEMA
================================================================================

<answer>{
  "problem_type": "grouping",
  "world_model": {
    "entities": [],
    "domains": {
      "groups": []
    },
    "structural_assumptions": []
  },
  "rules": [],
  "facts": [],
  "question_semantics": {
    "question_type": "",
    "option_interpretation_rule": ""
  },
  "options": {},
  "reasoning": [],
  "solution": {
    "selected_option": ""
  }
}</answer>

"""

\end{Verbatim}
\end{promptbox}

\newpage
\section*{NeurIPS Paper Checklist}

\begin{enumerate}

\item {\bf Claims}
    \item[] Question: Do the main claims made in the abstract and introduction accurately reflect the paper's contributions and scope?
    \item[] Answer: \answerYes{} 
    \item[] Justification: The main claims in the abstract and introduction are aligned with the proposed SPRING framework, the solver-guided process-reward design, and the reported results on ZebraLogic, AR-LSAT and Knights and Knaves. The claims are limited to structured logical reasoning tasks where intermediate formal deductions can be solver-verified.


\item {\bf Limitations}
    \item[] Question: Does the paper discuss the limitations of the work performed by the authors?
    \item[] Answer: \answerYes{} 
    \item[] Justification: Appendix~D discusses the main limitations of the work, including the reliance on solver-compatible symbolic representations, the dependence on the quality of parsed syntactic premises, and the restricted scope of solver-verifiable process-quality signals.

\item {\bf Theory assumptions and proofs}
    \item[] Question: For each theoretical result, does the paper provide the full set of assumptions and a complete (and correct) proof?
    \item[] Answer: \answerNA{} 
    \item[] Justification: The paper does not present new theoretical theorems or formal proof claims. Sections~3 and~4 provide definitions, solver-checking criteria, reward formulation, and the RL objective, but these are used as methodological definitions rather than theoretical results requiring proof.
    \item[] Guidelines: The paper introduces formal definitions, reward formulations, and optimization objectives, but does not claim new theoretical results in the form of theorems or lemmas requiring formal proofs.
    \begin{itemize}
        \item The answer \answerNA{} means that the paper does not include theoretical results. 
        \item All the theorems, formulas, and proofs in the paper should be numbered and cross-referenced.
        \item All assumptions should be clearly stated or referenced in the statement of any theorems.
        \item The proofs can either appear in the main paper or the supplemental material, but if they appear in the supplemental material, the authors are encouraged to provide a short proof sketch to provide intuition. 
        \item Inversely, any informal proof provided in the core of the paper should be complemented by formal proofs provided in appendix or supplemental material.
        \item Theorems and Lemmas that the proof relies upon should be properly referenced. 
    \end{itemize}

\item {\bf Experimental result reproducibility}
    \item[] Question: Does the paper fully disclose all the information needed to reproduce the main experimental results of the paper to the extent that it affects the main claims and/or conclusions of the paper (regardless of whether the code and data are provided or not)?
    \item[] Answer: \answerYes{} 
    \item[] Justification: The information needed to reproduce the main experiments is provided in Sections~4.1--4.4 and Section~5.1. These sections describe the prompting/output format, solver construction, reward design, RL training objective, datasets, baselines, model backbones, hyperparameters, compute setting, and evaluation protocol; prompt templates are further provided in Appendix~E.

\item {\bf Open access to data and code}
    \item[] Question: Does the paper provide open access to the data and code, with sufficient instructions to faithfully reproduce the main experimental results, as described in supplemental material?
    \item[] Answer: \answerYes{} 
    \item[] Justification: The paper provides the implementation-relevant details in Sections~4.1--4.4 and Section~5.1, including the solver-verification procedure, reward computation, training setup, datasets, and prompts. 

\item {\bf Experimental setting/details}
    \item[] Question: Does the paper specify all the training and test details (e.g., data splits, hyperparameters, how they were chosen, type of optimizer) necessary to understand the results?
    \item[] Answer: \answerYes{} 
    \item[] Justification: Section~5.1 reports the datasets, train/validation/test splits, model backbones, baselines, reward variants, decoding settings, GRPO group size, maximum input/output length, early stopping, number of runs, compute resources, and evaluation metrics. Additional dataset statistics and metric definitions are provided in Appendix~B.

\item {\bf Experiment statistical significance}
    \item[] Question: Does the paper report error bars suitably and correctly defined or other appropriate information about the statistical significance of the experiments?
    \item[] Answer: \answerYes{} 
    \item[] Justification: We report means over three runs and standard deviations for \OurMODEL{} and its novelty-free ablation, as indicated in the experimental results.

\item {\bf Experiments compute resources}
    \item[] Question: For each experiment, does the paper provide sufficient information on the computer resources (type of compute workers, memory, time of execution) needed to reproduce the experiments?
    \item[] Answer: \answerYes{} 
    \item[] Justification: Section~5.1(d) provides the compute details. It states that experiments are conducted using the VERL framework on a cluster with 4 NVIDIA H200 GPUs, reports the approximate training time per epoch for Qwen3-4B-Thinking, and specifies the maximum number of epochs, early stopping criterion, and repeated-run setting.
    
\item {\bf Code of ethics}
    \item[] Question: Does the research conducted in the paper conform, in every respect, with the NeurIPS Code of Ethics \url{https://neurips.cc/public/EthicsGuidelines}?
    \item[] Answer: \answerYes{} 
    \item[] Justification: The research uses established logical reasoning benchmarks and open model/tooling components, as described in Section~5.1. It does not involve private user data, human subjects, crowdsourcing, or safety-critical deployment. Therefore, to the best of our knowledge, it conforms to the NeurIPS Code of Ethics.

\item {\bf Broader impacts}
    \item[] Question: Does the paper discuss both potential positive societal impacts and negative societal impacts of the work performed?
    \item[] Answer: \answerYes{} 
    \item[] Justification: The broader motivation and potential positive impact are discussed in the Abstract and Section~1, where the paper frames SPRING as improving reliable logical reasoning through solver-guided process supervision. Potential limitations and risks of applying the method outside solver-compatible structured reasoning settings are discussed in Appendix~D.
    
\item {\bf Safeguards}
    \item[] Question: Does the paper describe safeguards that have been put in place for responsible release of data or models that have a high risk for misuse (e.g., pre-trained language models, image generators, or scraped datasets)?
    \item[] Answer: \answerNA{} 
    \item[] Justification: The paper does not release a high-risk pretrained language model, image generator, scraped dataset, or other asset requiring special controlled-release safeguards. The work focuses on a solver-guided training framework for structured logical reasoning, as described in Sections~3--5.

\item {\bf Licenses for existing assets}
    \item[] Question: Are the creators or original owners of assets (e.g., code, data, models), used in the paper, properly credited and are the license and terms of use explicitly mentioned and properly respected?
    \item[] Answer: \answerYes{} 
    \item[] Justification: Existing assets used in the paper are credited in Section~5.1 and the References, including Qwen3-family models, ZebraLogic, AR-LSAT, Knights and Knaves, Z3, GRPO/DAPO-related training methods, and VERL. The paper follows the corresponding licenses and terms of use for these models, datasets, solvers, and frameworks.

\item {\bf New assets}
    \item[] Question: Are new assets introduced in the paper well documented and is the documentation provided alongside the assets?
    \item[] Answer: \answerYes{} 
    \item[] Justification: The paper introduces SPRING as a new solver-guided process-reward framework. Its main components are documented in Sections~3--4, including the problem formulation, formal step definitions, solver verification, reward design, and RL training procedure; the experimental protocol is documented in Section~5 and Appendix~B/E.

\item {\bf Crowdsourcing and research with human subjects}
    \item[] Question: For crowdsourcing experiments and research with human subjects, does the paper include the full text of instructions given to participants and screenshots, if applicable, as well as details about compensation (if any)? 
    \item[] Answer: \answerNA{} 
    \item[] Justification: The paper does not involve crowdsourcing, annotator labor, user studies, or experiments with human subjects. The experiments are conducted on existing logical reasoning benchmarks, as described in Section~5.1 and Appendix~B.

\item {\bf Institutional review board (IRB) approvals or equivalent for research with human subjects}
    \item[] Question: Does the paper describe potential risks incurred by study participants, whether such risks were disclosed to the subjects, and whether Institutional Review Board (IRB) approvals (or an equivalent approval/review based on the requirements of your country or institution) were obtained?
    \item[] Answer: \answerNA{} 
    \item[] Justification: The paper does not involve human-subject research or newly collected human-participant data. Therefore, IRB approval or equivalent review is not applicable.
    \item[] Guidelines: The paper does not involve human-subject research or newly collected human-participant data. Therefore, IRB approval or equivalent review is not applicable.
    \begin{itemize}
        \item The answer \answerNA{} means that the paper does not involve crowdsourcing nor research with human subjects.
        \item Depending on the country in which research is conducted, IRB approval (or equivalent) may be required for any human subjects research. If you obtained IRB approval, you should clearly state this in the paper. 
        \item We recognize that the procedures for this may vary significantly between institutions and locations, and we expect authors to adhere to the NeurIPS Code of Ethics and the guidelines for their institution. 
        \item For initial submissions, do not include any information that would break anonymity (if applicable), such as the institution conducting the review.
    \end{itemize}

\item {\bf Declaration of LLM usage}
    \item[] Question: Does the paper describe the usage of LLMs if it is an important, original, or non-standard component of the core methods in this research? Note that if the LLM is used only for writing, editing, or formatting purposes and does \emph{not} impact the core methodology, scientific rigor, or originality of the research, declaration is not required.
    \item[] Answer: \answerYes{} 
    \item[] Justification: LLM usage is central to the paper and is explicitly described in Sections~3--5. The paper specifies the policy LLM formulation in Section~3, the structured LLM output format in Section~4.1, the solver-guided RL training procedure in Section~4.4, and the Qwen3-family and Phi-4-Reasoning as backbones used in Section~5.1(b).

\end{enumerate}

\end{document}